\documentclass{article} 
\usepackage{iclr2025_conference,times}

\usepackage{amsmath,amsfonts,bm}

\def\eqref#1{equation~\ref{#1}}

\def\1{\bm{1}}

\DeclareMathAlphabet{\mathsfit}{\encodingdefault}{\sfdefault}{m}{sl}
\SetMathAlphabet{\mathsfit}{bold}{\encodingdefault}{\sfdefault}{bx}{n}

\usepackage[utf8]{inputenc} 
\usepackage[T1]{fontenc}    
\usepackage{hyperref}       
\usepackage{url}            
\usepackage{subcaption}
\usepackage{dirtytalk}
\usepackage{wrapfig}
\usepackage{listings}
\usepackage{graphicx} 
\usepackage{enumitem}
\setlist[itemize]{leftmargin=*}
\usepackage{xcolor}
\usepackage{tikz}
\usepackage[most]{tcolorbox}
\usepackage[margin=1in]{geometry}
\usepackage{kotex}
\usepackage{amsmath, algorithm, algpseudocode}
\usepackage{ulem}
\usepackage{fontawesome}
\usepackage{float}
\usepackage{booktabs}

\usepackage{hyperref}
\usepackage{blindtext}
\newcommand{\dimension}[5]{
    \begin{tcolorbox}[skin=bicolor,fonttitle=\bfseries,coltitle=black,colbacktitle=#2!50,colback=#2!20,colframe=#2,title=#1, after skip=0.35em,left=3pt, right=3pt, top=3pt, bottom=3pt,boxsep=0pt,colbacklower=#2!10,middle=0.4em,toptitle=6pt, bottomtitle=4pt,sharp corners=all,boxrule=0mm, leftrule=1mm, title=#3]
    #4
    \end{tcolorbox}%
    \vspace{1em}\noindent#5
}

\usepackage{color,soul}

\usepackage{subcaption}
\definecolor{accuracy}{HTML}{A64D78}
\definecolor{uncertainty}{HTML}{765FAF}
\definecolor{attention}{HTML}{A0C5E8}

\makeatletter
\newtcbox{\kwColorBox}[1][]{on line,fontupper=\footnotesize\sffamily\bfseries\small,boxrule=1.5pt,arc=2pt,coltext=gray,colback=#1!10!white,colframe=#1,boxsep=0pt,left=1.5pt,right=1.5pt,top=1.5pt,bottom=1.5pt}
\makeatother

\newcommand{\refRO}[1]{\hyperref[challenge:#1]{\rawchallengedonotuse{#1}}}

\title{\textbf{Why context matters in VQA \& Reasoning}:\\ Semantic interventions for VLM input modalities}

\author{Kenza Amara \textsuperscript{\textbf{*}} $^{1,2}$, Lukas Klein \textsuperscript{\textbf{*}} $^{1,3,4}$, Carsten Lüth$^{3,4}$, Paul Jäger$^{3,4}$, Hendrik Strobelt$^5$, Mennatallah El-Assady$^1$ \\
$^1$ ETH Zurich, $^2$ ETH AI Center,
$^3$ German Cancer Research Center (DKFZ),\\
$^4$ Helmholtz Imaging,
$^5$ IBM Research\\
\texttt{kenza.amara@ai.ethz.ch, lukas.klein@dkfz.de}
}

\usepackage[symbol]{footmisc}

\iclrfinalcopy 
\begin{document}

\maketitle

\footnotetext[1]{Contributed Equally}

\begin{abstract}
The various limitations of Generative AI, such as hallucinations and model failures, have made it crucial to understand the role of different modalities in Visual Language Model (VLM) predictions. Our work investigates how the integration of information from image and text modalities influences the performance and behavior of VLMs in visual question answering (VQA) and reasoning tasks. We measure this effect through answer accuracy, reasoning quality, model uncertainty, and modality relevance. We study the interplay between text and image modalities in different configurations where visual content is essential for solving the VQA task. Our contributions include (1) the Semantic Interventions (SI)-VQA dataset, (2) a benchmark study of various VLM architectures under different modality configurations, and (3) the Interactive Semantic Interventions (ISI) tool.
The SI-VQA dataset serves as the foundation for the benchmark, while the ISI tool provides an interface to test and apply semantic interventions in image and text inputs, enabling more fine-grained analysis. Our results show that complementary information between modalities improves answer and reasoning quality, while contradictory information harms model performance and confidence. Image text annotations have minimal impact on accuracy and uncertainty, slightly increasing image relevance. Attention analysis confirms the dominant role of image inputs over text in VQA tasks. In this study, we evaluate state-of-the-art VLMs that allow us to extract attention coefficients for each modality. A key finding is PaliGemma's harmful overconfidence, which poses a higher risk of silent failures compared to the LLaVA models. This work sets the foundation for rigorous analysis of modality integration, supported by datasets specifically designed for this purpose. The code is available at \url{https://gitlab.com/dekfsx1/si-vlm-benchmark} and the tool and dataset are hosted at \url{https://gitlab.com/dekfsx1/isi-vlm}.
%

%
\end{abstract}

\section{Introduction}

Vision-Language Models (VLMs) have shown remarkable performance across various NLP tasks by combining visual and textual information. 
Models like LLaVA \citep{liuVisualInstructionTuning2023a}, GPT4-Vision \citep{openaiGPT4TechnicalReport2024}, and PaliGemma \citep{beyerPaliGemmaVersatile3B2024} use the text interface of Large Language Models (LLMs) alongside CLIP-style image encoders \citep{radford2021learningtransferablevisualmodels}, making them well-suited for multi-modal tasks such as content summarization \citep{moured2024altchartenhancingvlmbasedchart}, text-guided object detection \citep{dorkenwald2024pinpositionalinsertunlocks}, and visual question answering (VQA) \citep{yue2023mmmu}.

The core principle of VLMs lies in the integration and interplay of the two modalities, which significantly amplifies the effectiveness of the models. 
However, the extent to which each modality influences the final answer and reasoning of VLMs remains unclear within the research community, with varying and sometimes conflicting conclusions~\citep{gat2021perceptual,frank2021vision,MMStar}.
Prior research has highlighted the dominance of textual data \citep{Parcalabescu_2023}, but such studies often operate in settings where each modality, i.e., image or text, could independently be used to solve a task. 
This definition of multimodality, focusing on two independent but informative inputs, limits our understanding of how these modalities interact. 
Establishing a dataset and a tool to benchmark and examine the role of input modalities in model predictions within a multimodal setting would therefore significantly benefit the community. 
This approach would not only aid in understanding and mitigating model failures, such as hallucinations but also guide researchers toward practical solutions for improving model performance and self-consistency.

To this end, we present a benchmark, investigating the effect of semantic interventions on text and image modalities, alongside a carefully curated dataset and an interactive tool.
Specifically, we investigate the role of the text modality as a context guide in VQA tasks. 
We make the image the necessary data source to answer the question while using the text in different configurations to affect the model's answer and reasoning. Initially, we develop the \textbf{Interactive Semantic Interventions (ISI) Tool} to perform inter-modality interventions, testing their effect on the model behavior and observing which interventions lead to model failure. Based on the results, we curate the \textbf{Semantic Interventions (SI-)VQA dataset}. 
It contains 100 cautiously crafted examples with controlled interventions in both the input image and context. 
Each instance includes an image, the same image annotated, a complementary or contradictory textual context, and a question with a ground truth Yes/No answer, which can only be inferred from the image. 
This dataset allows for a wide range of image-text combinations. Using the SI-VQA Dataset, we establish a \textbf{comprehensive benchmark} to assess the impact of various interventions on VLMs across multiple evaluation criteria.
We experiment with different configurations where text serves as either complementary or contradictory context to the image, or as annotations directly on the image. 
We then analyze the model's performance, uncertainty, and attention attribution across these modalities.
Specifically, we examine the latest open-source VLMs including LLaVA 1.5 \citep{liuVisualInstructionTuning2023a}, LLaVA-Vicuna \citep{liuImprovedBaselinesVisual2024}, LLaVA-NeXT \citep{liu2024llavanext}, and PaliGemma \citep{beyerPaliGemmaVersatile3B2024}.

Our study demonstrates that integrating complementary contextual information into VQA models enhances both their answer accuracy and reasoning quality, while contradictory information significantly degrades performance and confidence, comparable to the absence of visual input. Additionally, we find that these models inherently prioritize visual data over textual context; efforts to rebalance attention toward textual information—such as adding image annotations, textual descriptions, or prompt engineering—yield mixed results, with prompt engineering improving accuracy without altering attention distribution and textual descriptions unexpectedly reducing accuracy and increasing uncertainty. Using the AUGRC score to measure the frequency of silent failures, our experiments reveal the harmful overconfidence of the PaliGemma 3B model compared to the LLaVA models. 
These findings provide valuable insights for the AI community by highlighting the critical role of context in VQA models, guiding future developments to optimize attention between visual and textual modalities, and assessing the risk of silent model failures.

In summary, our contributions are:

\begin{itemize}
    \item A large-scale benchmark to evaluate the semantic interplay between image and textual context based on diverse evaluation criteria and several state-of-the-art VLMs, meant to facilitate silent failure detection.
    \item Ablation studies regarding rebalancing attention to specific modalities and the effect of model quantization.
    \item The SI-VQA Dataset, a well-curated dataset that enables the exploration of various combinations of image and text modalities. The benchmark and dataset are hosted at: \url{https://gitlab.com/dekfsx1/si-vlm-benchmark}.
    \item The ISI Tool, a tool designed to enable users to investigate how VLMs respond to semantic changes and interventions across image and text modalities, with a focus on identifying potential model failures. The tool is hosted at: \url{https://gitlab.com/dekfsx1/isi-vlm}. 
\end{itemize}

\begin{figure}[t!]
    \centering
    \includegraphics[page=1, trim=0cm 0cm 0cm  0cm, clip, width=\linewidth]{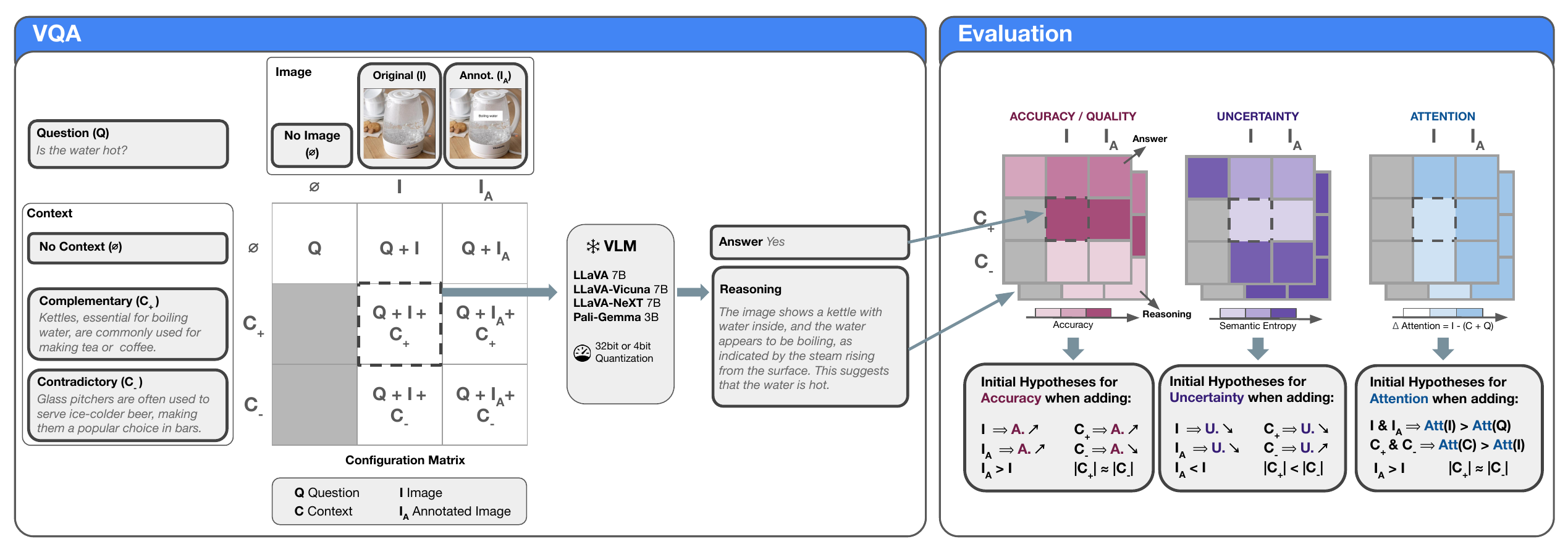}
     \caption{The SI-VQA framework examines the influence of various modality configurations on \textbf{\textcolor{accuracy}{answer accuracy and reasoning quality}}, model \textbf{\textcolor{uncertainty}{uncertainty}}, and \textbf{\textcolor{attention!160}{attention attribution}}. Seven different configurations are tested, combining inputs such as the question (Q), image (I), annotated image (I$_{\text{A}}$), and either complementary (C$_{+}$) or contradictory context (C$_{-}$): (Q), (Q+I), (Q+I+C$_{+}$), (Q+I+C$_{-}$), (Q+I$_{\text{A}}$), (Q+I$_{\text{A}}$+C$_{+}$), and (Q+I$_{\text{A}}$+C$_{-}$). For each configuration, the VLM is assessed first on its answer and then on its reasoning. Furthermore, we establish prior assumptions regarding how each modality is expected to impact the model's behavior.}
    \label{fig:overview}
\end{figure}

\section{Related Work}\label{sec:related_work}

\paragraph{VQA Datasets}
Since the introduction of the Visual Question Answering (VQA) task in \citet{antol2015vqa}, numerous datasets have been created to support research in this area, such as GQA \citep{hudson2019gqa}, Visual7W \citep{zhu2016visual7w}, Visual Genome \citep{krishna2017visual}, CLEVR \citep{johnson2017clevr}, and VQA 2.0 \citep{goyal2017making}. However, these datasets often lack reasoning, are restricted to textual-only modalities, or suffer from small-scale and limited domain diversity. ScienceQA \citep{ScienceQA} emerged as the first large-scale VQA dataset to incorporate textual context alongside textual explanations for model reasoning. It features multiple-choice questions across various scientific domains, with each question annotated with lectures and explanations. More diverse multimodal datasets, such as SEED \citep{SEED}, MMBench \citep{MMBench}, and MMMU \citep{MMMU}, have since been introduced, combining text and image data. However, as shown by \citet{MMStar}, these datasets do not ensure that all evaluation samples require visual content for correct answers.
To address this limitation, we propose a new curated VQA dataset designed such that answers can only be derived from the image, with complementary or contradictory context provided as additional information to influence the model’s answer and reasoning. 

\paragraph{Modalities in Vision-Language Tasks}

Several methods have been proposed to investigate the extent to which VLMs leverage both visual and textual information.
\textit{Annotation and foiling approaches} introduce text annotation in images and mistakes in image descriptions and test whether the VLM predictions change. \citet{shekhar-etal-2019-evaluating, parcalabescu-etal-2022-valse} test VLMs' sensitivity to discrepancies between images and captions and found that models often overlook such inconsistencies.
\citep{gat2021perceptual} exchange images and captions with other instances and note a consistent decrease in accuracy, with textual input proving to be more influential than visual content. Building on these findings, we incorporate textual annotations into images in our study to explore how textual data, shown in previous work to be more significant than pixels, can support model reasoning. 
\textit{Ablation methods} investigate model behavior when parts of the input are removed or masked~\citep{bugliarello-etal-2021-multimodal, cafagna2021vision}. \citet{frank2021vision} occludes parts of the image or masks the text, and finds that the visual modality matters more than the text. In line with this, we introduce semantic perturbations, creating scenarios where the text complements the image in various ways.
\textit{Attention-based methods} correlate high attention scores with high feature importance, though this connection remains debated. \citep{serrano-smith-2019-attention, jain-wallace-2019-attention} questioned the validity of attention as an indicator of importance, while \citep{wiegreffe-pinter-2019-attention} challenged their arguments, asserting that attention can serve as an explanation, albeit not the definitive one. Following mechanistic interpretability, our paper also analyzes how attention scores are attributed across modalities in different text-image configurations. We view attention attribution as one key factor in understanding the intrinsic roles of each modality in VLM performance.
Early work on the impact of modalities in VQA has produced mixed results. 
While some studies argue that VLMs rely more on text \citep{gat2021perceptual},  others emphasize the importance of the visual modality \citep{frank2021vision}. More recently, \citet{MMStar} showed that visual data might be unnecessary to answer several questions, suggesting among others unintentional data leakage through LLM and VLM training. Our paper builds on this debate by proposing a rigorous framework for conducting semantic interventions on both modalities to assess their respective contributions to VQA tasks.

\paragraph{Multimodal XAI}
Multimodal Explainable AI (MXAI) is a branch of XAI that includes a range of XAI techniques tailored to address the unique challenges posed by multimodal data inputs, tasks, and architectures. Recent reviews study the new challenges and differences of those methods as compared to traditional XAI approaches~\citep{joshi2021review,rodis2023multimodal}.
In the multimodal context, MXAI leverages the complementary explanatory strengths of different modalities offering richer insights. In specific scenarios, language can provide a deeper understanding and clarification of concepts, while in others, the visual modality may be more informative~\citep{park2018multimodal}. MXAI is used as a tool to interpret VQA tasks that might require higher-order reasoning and a deep understanding of semantic context \citep{antol2015vqa}. 

\section{SI-VQA Dataset and ISI Tool}

This section introduces the dataset and tool developed to examine the role of modalities in VQA \& Reasoning. They are designed to explore semantic interventions on input modalities for thorough post-hoc interpretability of VLMs.

\subsection{Si-VQA Dataset}

The SI-VQA dataset is a closed-question VQA dataset consisting of 100 samples. Each sample consists of an image, question, and ground truth answer (Yes/No) pair, as well as one text annotated image, one contradictory context, and one complementary context. The image is the necessary and sufficient element to get the question right as it contains the critical information. The context is always in relation to the ground truth answer, aiming to either confuse or help the model without stating an explicit answer to the question, as each question can only be answered through the image. Thus, the model has a very limited ability to leverage prior knowledge to answer the question and must utilize the image input. The annotations on the image give text-written hints to the ground truth answer. We only study the impact of ``positive'' annotations, i.e., text that informs the model about key elements in the image to correctly answer the question. See \autoref{fig:overview} for an exemplary sample. The images are open-source and from the MMMU Benchmark \citep{yue2023mmmu}. The selected images of SI-VQA Dataset span a variety of fields, including geography, history, art and design, sport, and biology, along with everyday objects and landscapes. They encompass diverse formats, such as natural photographs, cartoons, sketches, and paintings. We carefully crafted all other modalities for the dataset, focusing on quality not quantity.

We design seven scenarios for VLM interpretability analysis, creating seven modality configurations: question (Q), question + image (Q+I), question + image + complementary context (Q+I+C$_{+}$), question + image + contradictory context (Q+I+C$_{-}$), question + annotated image (Q+I$_{\text{A}}$), question + annotated image + complementary context (Q+I$_{\text{A}}$+C$_{+}$), question + annotated image + contradictory context (Q+I$_{\text{A}}$+C$_{-}$). They can be inferred using the 3x3 matrix in~\autoref{fig:overview}. For the baseline configuration Q (question-only), the image corresponds to a black image. In this case, the model's answer accuracy is at random (See Appendix~\autoref{apx:vaqacc_results}) proving the necessity of visual content in SI-VQA Dataset~\citep{MMStar}. We also conducted the baseline experiment using noise-only images and with the images entirely removed, observing similar results in both cases (See \autoref{apx:vqa_baseline}).

\subsection{ISI Tool}

To provide an intuitive and agile way to explore modality interplay in the context of multimodal interpretability, we developed the ISI Tool.
We used this interactive tool to design the SI-VQA Dataset.
It is designed to enable researchers and VLM users to investigate how VLMs respond to semantic changes and interventions across image and text modalities, with a focus on identifying potential model failures in the context of VQA.
Specifically, it allows the perturbation of images, the addition of personalized shapes and annotations, and the arbitrary adaptation of text inputs.
Users can upload their own images and questions or choose from 100 preloaded samples with semantic intervention presets from our ISI-VQA Dataset. 
More details about the tool and its features are available in \autoref{apx:isitool}.

\section{Experiment Methodology}

Our benchmark pipeline illustrated in~\autoref{fig:overview} investigates the contribution of each modality toward the performance of several VLMs. Performance is measured in terms of output quality, model uncertainty, and attention attribution toward the input elements of the SI-VQA Dataset.

\subsection{Vision-Language Models} \label{subsec:models}

The VLMs selected for this study are state-of-the-art models for Visual Question Answering (VQA) tasks. We excluded models where extracting attention coefficients for each modality is not feasible, such as Flamingo~\citep{alayrac2022flamingo}, which employs gated cross-attention between text and image. The final architectures chosen for evaluation are LLaVA 1.5, LLaVA-Vicuna, LLaVA-NeXT, and PaliGemma~\citep{beyerPaliGemmaVersatile3B2024}, with weights provided from HuggingFace \citep{wolf2020huggingfacestransformersstateoftheartnatural}. LLaVA-Vicuna is a version of LLaVA 1.5 leveraging the Vicuna LLM, a conversation-fine-tuned version of LLaMA. LLaVA-Vicuna and LLaVA-NeXT both utilize dynamic high resolution for the image input \citep{liuImprovedBaselinesVisual2024}, increasing visual reasoning and optical character recognition (OCR) capabilities. 
Although PaliGemma is intentionally designed for pre-training followed by fine-tuning, we employ it in this study within a zero-shot setting.
We do not present any reasoning results for the PaliGemma model as it mainly generates a default response ``Sorry, as a base VLM I am not trained to answer this question.'' when it is asked to explain its answer due to its strong safety fine-tuning~\citep{beyerPaliGemmaVersatile3B2024}.
Each LLaVA architecture comprises 7B parameters, while PaliGemma consists of 3B parameters, and all can be quantized to reduce VRAM usage and improve computational efficiency.
Our ablation study in \autoref{apx:4bit} shows, however, that while 32bit models have lower uncertainty in VQ answering and reasoning, answer accuracy is not substantially worse for even 4bit quantized models.

\subsection{Answer \& Reasoning Evaluation}

We assess the model’s performance on the VQA task using the SI-VQA Dataset by evaluating both its answer and reasoning. To measure the VQ answering quality, we use the accuracy metric by comparing the model's binary Yes/No response to the closed question with the ground truth provided in the dataset. Evaluating reasoning is more complex, as no ground truth exists for the explanations. Without a reference rationale, we assess reasoning based on the quality of argumentation and the truthfulness of statements. To this end, we useo as an external evaluator \citep{zhengjudging2024,thakur2024judging}. The model is prompted once for each sample to rate the reasoning from 0 to 10, considering an evaluation prompt as well as the question, image, answer, and reasoning. While the quality scores seem very reasonable to us, we observe a bias toward the score number "8" (see \autoref{apx:vaqacc_results}), a behavior also observed in other studies using LLMs as a judge \citep{thakur2024judging}. See \autoref{apx:hyperparameters} for the evaluation prompts and hyperparameters.

\subsection{Model Uncertainty}
\begin{equation} \label{eq:se}
    SE(x) = - \sum_c p(c | x) \log p(c|x) = - \sum_c \left( \Big( \sum_{s\in c} p(s|x) \Big) \log \Big[ \sum_{s\in c} p(s|x) \Big] \right)
\end{equation}\\
For quantifying model uncertainty, we employ semantic entropy $SE()$ \citep{farquharDetectingHallucinationsLarge2024}, which calculates entropy based on the sum of token likelihoods $p()$ between the sets $c$ of semantically similar clustered sentences $s$ (see \autoref{eq:se}). For semantic clustering, we use the DeBERTa~\citep{he2021deberta} entailment model. During uncertainty computation, the number of sampled outputs and the sampling temperature $T$ are set to 10 and 0.9 respectively. High $SE()$ means high uncertainty and low confidence in the outputs.

For VLMs, uncertainty quantification is especially important for detecting model failures, including hallucinations and silent failures. 
Silent failures are instances where the model generates incorrect information with high confidence, making these errors increasingly difficult to detect \citep{bender2021dangers,jaeger2023reflectevaluationpracticesfailure}.
The Area Under the Generalized Risk Coverage curve (AUGRC) metric \citep{traub2024overcomingcommonflawsevaluation} evaluates the extent to which a model makes incorrect predictions with high confidence, where high confidence is characterized by low semantic entropy, specifically below a defined threshold, $\tau$:
\begin{equation}\label{eq:augrc}
    \ensuremath{\mathrm{AUGRC}}\xspace = \int_{0}^{1}P(Y_\text{f}=1,\,SE(x)\leq\tau)\,\mathrm{d}P(SE(x)\leq\tau) \quad
\end{equation}\\
With $Y_\text{f}$ as the binary failure indicator, i.e., $Y_\text{f}=1$ indicates a wrong prediction by the model. We refer to \citet{traub2024overcomingcommonflawsevaluation} for a detailed explanation. The AUGRC therefore directly measures the harmful overconfidence of a model, i.e., when the model is confident and wrong. A high AUGRC indicates that the model has a tendency for silent failures.
 
\subsection{Attention Attribution}

Attributing attention values to the different modalities serves as one indicator of each modality's contribution to the final answer and reasoning~\citep{wiegreffe-pinter-2019-attention}. We adopt a mechanistic approach, treating attention as a VLM interpretability measure to evaluate the roles of the question, image, and context. By aggregating attention across layers and heads, we derive relevance scores for each token's contribution to the answer or reasoning \citep{Parcalabescu_2023,parcalabescu2024measuring}. 
To determine which input modality is most relevant to the prediction, we sum the relevance scores of their respective input tokens. 
Finally, to standardize the scores for comparability, we compute the relative relevance score for each sample.
A detailed explanation of attention aggregation and implementation is provided in~\autoref{apx:attention}.

\section{Experiment Results}

\subsection{Initial Hypotheses} \label{sec:inithypo}

Based on the seven modality configurations and the results of previous related work, we define three prior hypotheses regarding the anticipated outcomes when including a new type of input. 
As discussed in \autoref{subsec:models}, only the VQ answering results are shown for PaliGemma.


\textbf{\textit{Hypothesis 1 (Including Image):}}
In the SI-VQA dataset, images are essential for VQA tasks, as they serve as the only modality providing the necessary information to answer the questions. 
Therefore, we hypothesize a significant positive impact on accuracy and a reduction in model uncertainty when the image is incorporated alongside the question. 
Furthermore, the natural image is expected to receive greater attention compared to the baseline black or random pixel images in the question-only configuration. 
The impact on reasoning quality, however, remains in our opinion uncertain.

\textbf{\textit{Hypothesis 2  (Including Context):}} 
We hypothesize that adding complementary context will improve model accuracy, confidence, and reasoning quality. 
This additional information is expected to facilitate more detailed and precise rationalization. 
Conversely, contradictory context is anticipated to decrease model accuracy, confidence, and reasoning quality, reflecting the model's doubt. 
Based on previous research, we also expect the model to exhibit a higher attention toward text compared to the image.

\textbf{\textit{Hypothesis 3 (Including Image Annotations):}} 
Compared to configurations including the natural image, those utilizing the annotated image are expected to enhance the model's ability to answer and reason with greater accuracy and confidence. 
Additionally, we anticipate increased attention toward the annotated image relative to the natural image.

In the following sections, we will first present the benchmark results before comparing them to the prior hypotheses we have just put forward.

\subsection{Answer \& Reasoning Evaluation}

\begin{figure}[ht!]
    \centering
    \includegraphics[page=1, trim=0cm 95cm 0cm  0cm, clip, width=\linewidth]{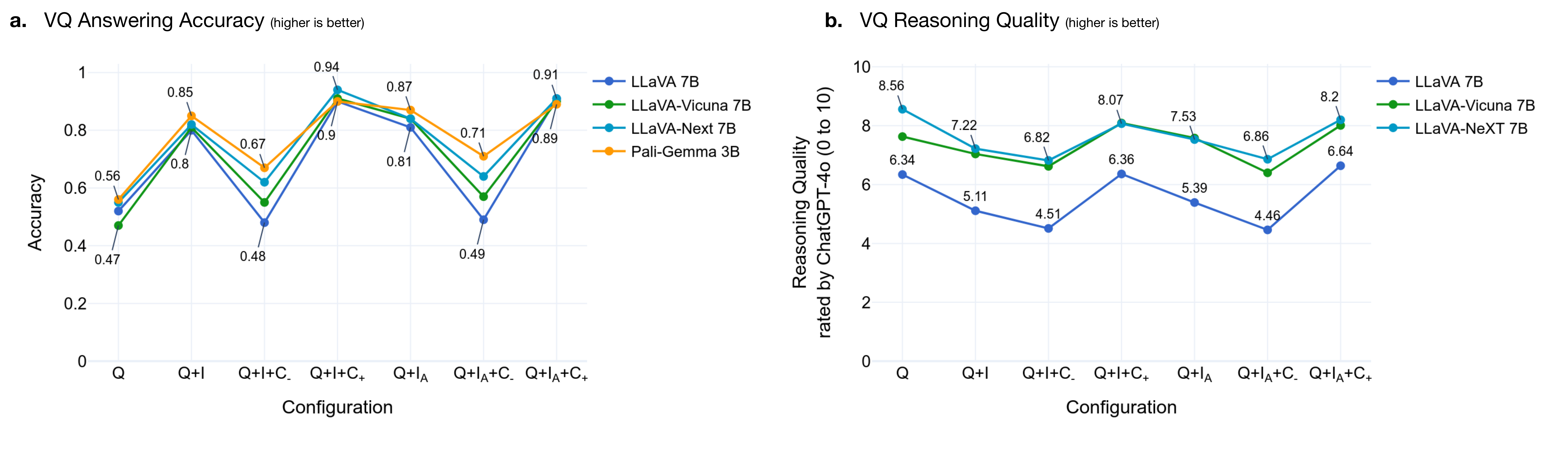}
    \vspace{-3em}
     \caption{Quality of VLM answers and reasoning in the seven modality configurations of question (Q), image (I), annotated image (I$_A$), complementary (C$_{+}$) and contradictory context (C$_{-}$). Answer accuracy is measured using the ground truth labels of our SI-VQA Dataset and reasoning quality is evaluated using the external scoring of GPT-4o as a judge. A significant drop of accuracy in the answer and reasoning is observed for all models when adding contradictory context, i.e., Q+I+C$_{-}$ and Q+I$_A$+C$_{-}$. Results for PaliGemma 3B are only displayed for answering (see~\autoref{subsec:models}).}
    \label{fig:vqa_acc}
\end{figure}

\autoref{fig:vqa_acc} presents the results of the VQA accuracy (a.) and the quality of the VLM VQ reasoning (b.), as judged by GPT-4o. Similar patterns emerge across all models. The answer accuracy is low in the question-only baseline, where the model lacks sufficient information to provide correct answers. However, when given just a question and a black image, the models consistently offer strong reasoning quality, consistently justifying their response by acknowledging the absence of image information. 
Incorporating complementary context into the I+Q configuration enhances both answer accuracy and reasoning quality by providing additional details necessary for a correct response and a well-supported rationale. 
In contrast, the introduction of contradictory context significantly degrades response accuracy. The decline in accuracy is the smallest for PaliGemma, whereas for LLaVA, it drops to a level comparable to the question-only configuration. 
Additionally, reasoning quality declines as the models are misled by the conflicting information.
When exchanging the natural image with an annotated image, we observe no change in accuracy or reasoning quality, even for architectures optimized for OCR.

When comparing the VLM architectures, a notable discrepancy emerges in their handling of contradictory information, with models responding differently to contradictions (configuration Q+I+C$_{-}$). 
Surprisingly, PaliGemma demonstrates the most robustness in managing contradictions and achieving the highest accuracy scores in five out of seven configurations, despite having less than half the parameters of the LLaVA models and not being explicitly fine-tuned for VQA tasks. 
LLaVA-NeXT ranks second in accuracy but does not fully leverage its enhanced OCR capabilities when the annotated image is included. 
In terms of reasoning abilities, the conversational fine-tuned VLMs produce substantially higher-quality reasoning compared to the standard LLaVA 1.5 model.

\dimension{}{accuracy}{Answer Accuracy \& Reasoning Quality}{While answer accuracy is low in the question-only baseline, models still provide strong reasoning quality by acknowledging missing image data. Complementary context improves accuracy and reasoning quality, but contradictory context significantly degrades performance. We observe the strongest decline for LLaVA and smallest in PaliGemma. Replacing the natural image with an annotated one shows no effect on accuracy or reasoning quality.}

\subsection{Model Uncertainty} 

\begin{figure}[ht!]
    \centering
    \includegraphics[page=2, trim=0cm 95cm 0cm  0cm, clip, width=\linewidth]{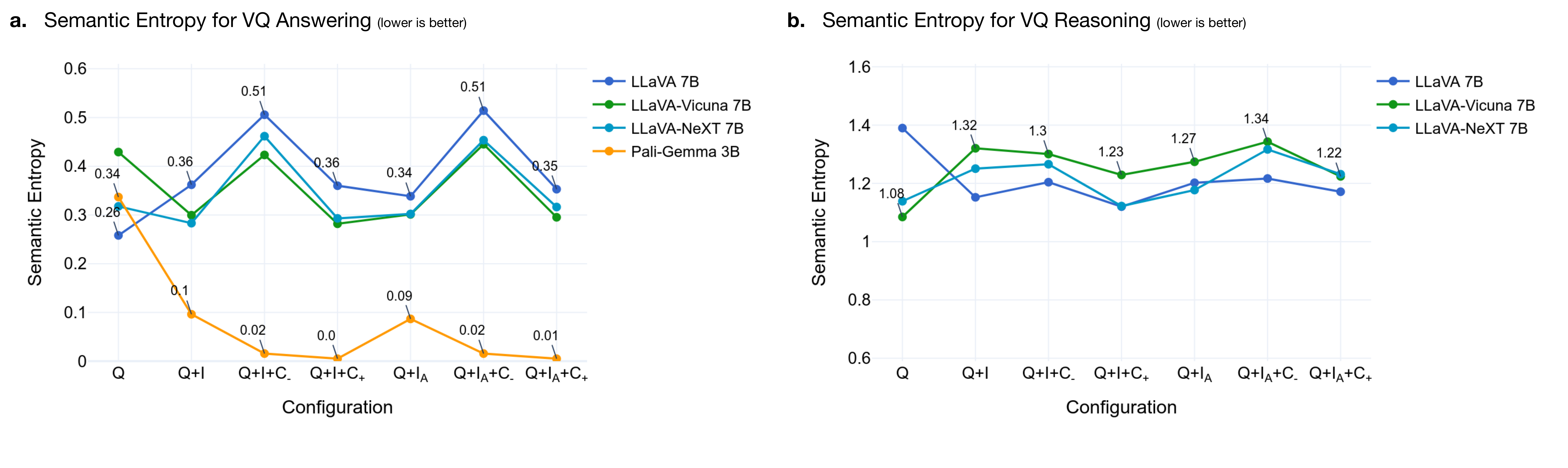}
    \vspace{-3em}
     \caption{VLM uncertainty when generating answers and reasonings in the seven modality configurations of question (Q), image (I), annotated image (I$_A$), complementary (C$_{+}$) and contradictory context (C$_{-}$). Uncertainty is measured using the semantic entropy--the lower the entropy, the more confident the model. PaliGemma 3B shows extreme confidence overall in its answers. However, no reasoning results for PaliGemma are provided (see~\autoref{subsec:models}). C$_{-}$ negatively impacts the certainty of LLaVA models when generating answers.}
    \label{fig:vqa_uncertainty}
\end{figure}

\autoref{fig:vqa_uncertainty} displays the model uncertainty measured through semantic entropy for both the answer (a.) and the reasoning (b.). For all models, the absence of image-based information (configuration Q) results in similar levels of uncertainty in both VQ answering and reasoning. In addition, we observe for all models that image text annotations have almost no impact on the model uncertainty compared to the configurations including the natural images.

\begin{wrapfigure}{r}{0.40\textwidth}
\vspace{.2cm}
    \centering
    \includegraphics[page=12, trim=0cm 104cm 98cm 0cm,clip,width=\linewidth]{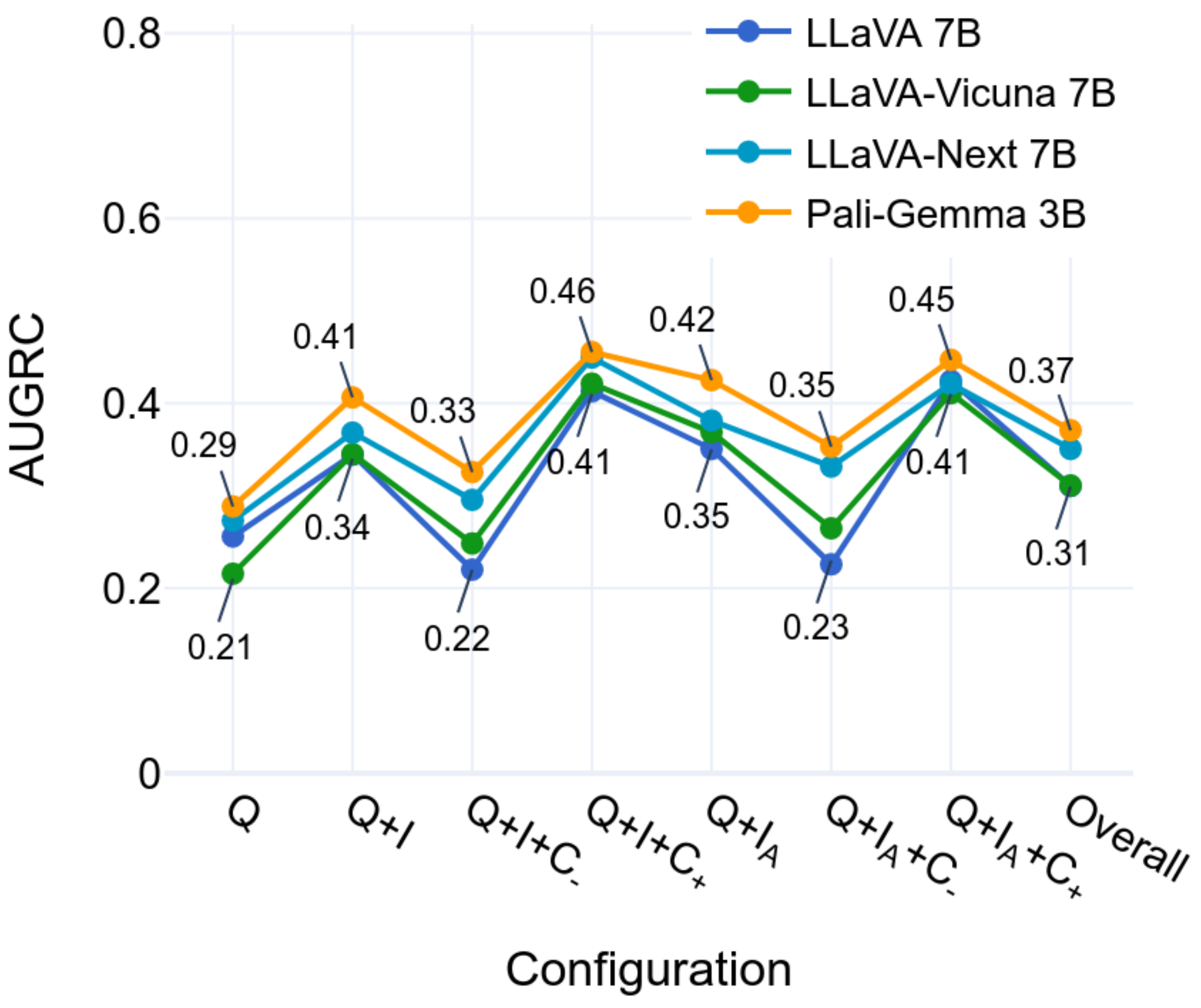}
  \caption{AUGRC evaluating the ability to detect silent failures through semantic entropy for each model and configuration (lower is better).}
  \label{fig:failure}
  \vspace{-.4cm}
\end{wrapfigure}
For PaliGemma, adding image and context information significantly reduces uncertainty in VQ answering, making the model much more confident in its predictions. It seems that providing additional context, regardless of its content, leads the model to be more self-assured. This intriguing pattern shows large overconfidence in PaliGemma, which does not always have to be beneficial as it can, e.g., lead to silent failures, where the model is extremely confident in its wrong predictions~\citep{bender2021dangers,jaeger2023reflectevaluationpracticesfailure}.

For all LLaVA models, we observe overall an inverse relationship between answer uncertainty and reasoning uncertainty, with LLaVA 1.5 exhibiting the highest uncertainty in VQ answering but the lowest in VQ reasoning. 

When the image is added, LLaVA-Vicuna and LLaVA-NeXT show reduced uncertainty in VQ answering but increased uncertainty in VQ reasoning, as the models, in the question-only configuration, only acknowledge the absence of the image and therefore reason with high confidence. Complementary context slightly decreases model uncertainty, indicating a marginal increase in confidence for both VQ answering and reasoning. This effect is minor though, as shown in~\autoref{fig:vqa_uncertainty} b. where all LLaVA models exhibit nearly identical semantic entropies for I+Q and Q+I+C${+}$, as well as for Q+I${_\text{A}}$ and Q+I${_\text{A}}$+C${+}$. Contradictory contextual information, on the other hand, significantly increases uncertainty in the model answers. Its effect on reasoning is also particularly pronounced in LLaVA 1.5 but remains relatively minor for LLaVA-Vicuna and LLaVA-NeXT. Thus, the LLaVA models appear to be slightly influenced by reinforcing information sources but are more easily unsettled by contradictory ones.

\paragraph{Model Failure Detection through Semantic Entropy}

Interpretating VLMs' behavior through the lense of model uncertainty is crucial for identifying and understanding failures, including hallucinations and silent failures. Specifically, for PaliGemma, silent failures cannot be dismissed due to the model's extreme overconfidence, despite its prediction accuracy being comparable to that of LLaVA models. To quantify this harmful overconfidence—when the model confidently predicts incorrect answers with very low uncertainty—we employ the AUGRC metric, as detailed in~\autoref{eq:augrc}. \autoref{fig:failure} displays the AUGRC values across all model architectures (where lower is better). Our results confirm that PaliGemma performs the worst, validating our hypothesis regarding its harmful overconfidence. Additionally, we observe that in cases of high uncertainty, such as with the Q+I+C$_{-}$ and Q+I$_\text{A}$+C$_{-}$ configurations, AUGRC is low, indicating fewer silent failures. In these scenarios, the contradictory context reduces the likelihood of confident incorrect answers, meaning the models become more uncertain about their mistakes, thereby making them more trustworthy.\\

\dimension{}{uncertainty}{Uncertainty}{
PaliGemma demonstrates high overconfidence, leading to silent failures, as evidenced by the AUGRC metric. LLaVA models display an inverse relationship between VQ answering and reasoning uncertainty, probably due to more detailed reasoning in more advanced models. Contradictory context significantly increases the VQ answering uncertainty for LLaVA models but has only a minor effect on VQ reasoning.}

\subsection{Attention Attribution}

\begin{figure}[ht!]
    \centering
    \includegraphics[page=3, trim=0cm 95cm 0cm  0cm, clip, width=\linewidth]{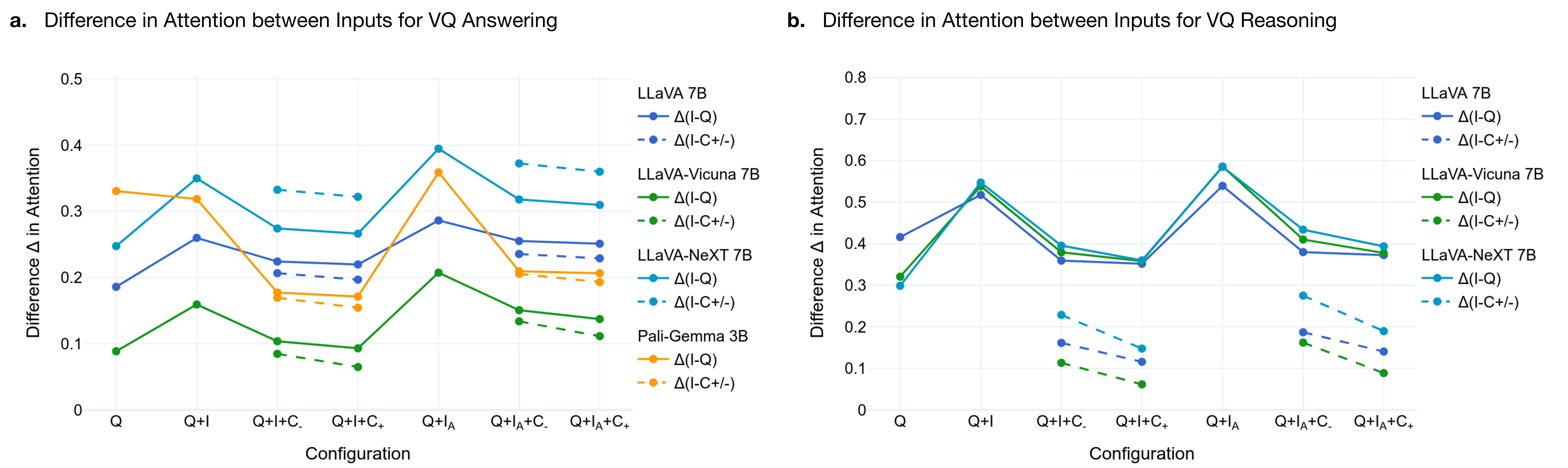}
    \vspace{-1em}
     \caption{Difference in attention attribution between the image and the question (solid line) and between the image and the context when present (dashed line). The computation of modality attention attribution is described in~\autoref{apx:attention}. The image (I) always gets the highest attention attribution compared to text modalities (Q, C$_{-}$, C$_{+}$). No reasoning results for PaliGemma are provided (see~\autoref{subsec:models}).}
    \label{fig:vqa_attention}
\end{figure}

This section examines how attention is distributed across the three inputs—question, image, and context—across the seven configurations. \autoref{fig:vqa_attention} shows the average difference between attention to the image and the question, as well as the attention to the image and the context.
Since all differences are positive, the image consistently receives the highest average attention in both VQ answering (\autoref{fig:vqa_attention}a.) and reasoning (\autoref{fig:vqa_attention}b.). 
The attention distribution across different inputs is similar among the models, with LLaVA-Vicuna showing the highest attention to textual inputs and LLaVA-NeXT focusing more on the image. 
Both answering and reasoning exhibit higher attention to the natural image compared to the black baseline image in the question-only configuration. 
Further, attention to the image decreases when context is added and the annotated image receives more attention than the natural image. 
In VQ answering, attention to the question and context is nearly equal, whereas, in VQ reasoning, the model shows significantly higher attention to the context, almost equal to the attention given to the image. 
Detailed figures are provided in Appendix \autoref{apx:attention_results}.
Overall, no strong correlation is observed between attention attribution and accuracy (see Appendix \autoref{apx:acc_corr}).

\dimension{}{attention}{Attention Attribution}{
The image modality receives the highest attention compared to the question and context, with its relevance further increasing when annotated. 
In VQ answering, attention to the question and context is nearly equal, whereas in VQ reasoning, the model allocates significantly more attention to the context.
LLaVA-Vicuna pays the highest attention to textual inputs and LLaVA-NeXT to the image.
}

\subsection{Rebalancing Modality Importance} \label{sec:rebalancing}

Given the observed high attention allocated to the image modality, it raises the question of how the results might change if we intervene to direct more attention toward the text modalities. 
Specifically, we aim to investigate how the model's performance changes when either the information in the image is described with text or the attention of the model is guided toward the text via prompt engineering.

\begin{figure}[ht!]
    \centering
    \includegraphics[page=4, trim=0cm 95cm 0cm  0cm, clip, width=\linewidth]{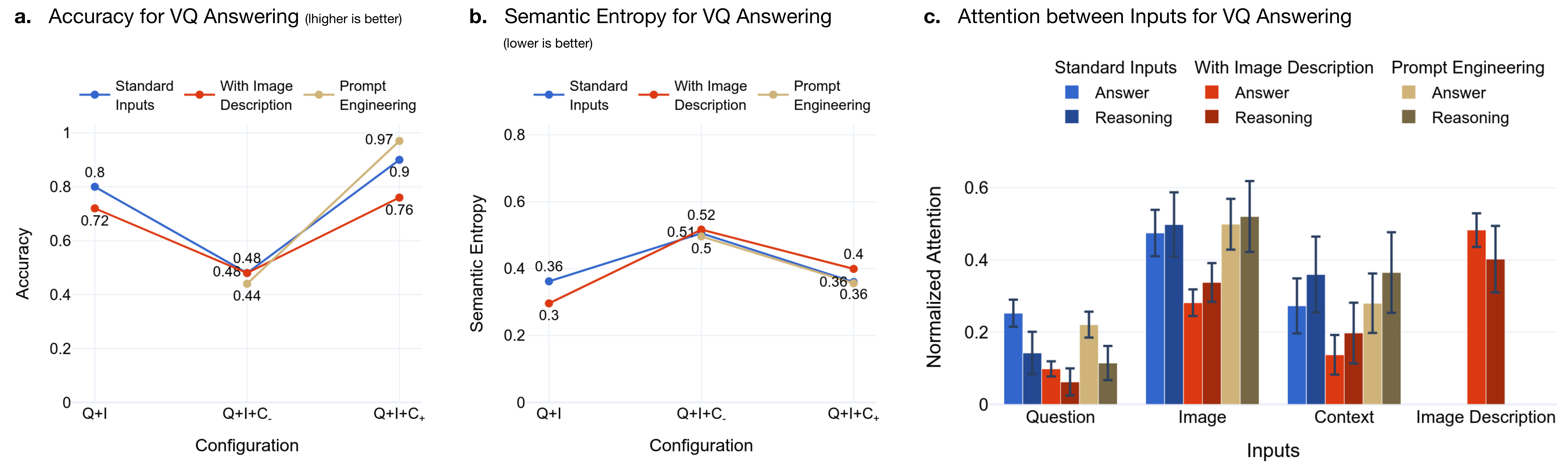}
 \caption{Two strategies to adjust modality importance: incorporating the image's textual description into the input and modifying the prompt to shift more attention toward the context. The impact of these changes is evaluated based on a. answer accuracy, b. model uncertainty, and c. attention attribution. The hereby experiments were conducted with the LLaVA 7B model, see \autoref{apx:rebalance_pali} for the PaliGemma results.}
    \label{fig:vqa_rebalance}
\end{figure}

\paragraph{Textual Description of the Image} \textit{
Does adding a text description of the image greatly improve model performance and confidence?} 
We initially hypothesized that augmenting the model's input with a textual description of the image would enhance its accuracy and reduce uncertainty, based on the premise that key features necessary for answering the question—already present in the image—would be more accessible to the LLM decoder in text form. 
However, as illustrated for LLaVA in \autoref{fig:vqa_rebalance}, the results reveal a surprising decrease in answer accuracy and an increase in model uncertainty for configuration I+Q+C$_+$.
We observe similar results for PaliGemma (see \autoref{apx:rebalance_pali}), with minor increases in uncertainty. 
We argue that these findings indicate the models are already proficient at extracting essential information from the image alone and that the addition of textual information introduces confusion due to redundancy or potential inconsistencies in the image description. 
Moreover, the high attention allocated to the image description in \autoref{fig:vqa_rebalance}c. underscores the model's sensitivity to textual inputs, which may inadvertently dominate visual cues.

\paragraph{Prompt Engineering} \textit{Can prompt engineering help VLMs re-balance their attention toward the context?}
We modified the initial prompt to direct the model's attention more toward the textual context, which typically receives less emphasis in the standard setting (see \autoref{apx:hyperparameters} for implementation details). 
Given that the complementary context provides information intended to guide the model toward the correct answer, while the contradictory context aims to mislead it, we expected an increase in accuracy for the Q+I+C$_{+}$ configuration and a decrease for the Q+I+C$_{-}$ configuration. 
As shown in \autoref{fig:vqa_rebalance} a, we observed a decrease in accuracy for Q+I+C$_{-}$, whereas prompt engineering to emphasize the complementary context resulted in improved answer accuracy. 
Unexpectedly, these changes are not reflected in the attention attribution: \autoref{fig:vqa_rebalance}c. indicates that prompt engineering does not alter the attention distribution across modalities, as the context does not receive increased attention. 
This lack of correlation between attention and model performance highlights the necessity for cautious interpretation of attention mechanisms in model predictions (see~\autoref{sec:related_work}).
Unlike the LLaVA results, PaliGemma exhibited a significant increase in uncertainty (see \autoref{apx:rebalance_pali}), highlighting the large influence of the image in the standard inputs. 

\section{Discussion \& Conclusion}

\paragraph{Evaluation of Hypotheses}
Our findings reveal notable insights into the role of each modality in VQA and reasoning tasks. 
Specifically, we compare all results with our initial hypotheses from \autoref{sec:inithypo}.

\textit{Hypothesis 1 (Including Image):}
As expected, introducing the image results in a significant increase in answer accuracy across all VLMs. 
However, it unexpectedly leads to a decrease in reasoning quality, as in the question-only setting, the models simply acknowledge the absence of the image. 
We observe a similar pattern in model uncertainty: it decreases for VQ answering but increases in reasoning. 
As hypothesized, the natural image indeed receives more attention compared to the black baseline image.

\textit{Hypothesis 2 (Including Context):}
Consistent with our expectations, the inclusion of complementary context enhances both accuracy and reasoning quality, while contradictory context has a strongly negative effect. 
However, in VQ answering, the complementary context does not reduce model uncertainty, whereas the contradictory context significantly increases it. 
In VQA reasoning, contradictory context does not affect uncertainty, and complementary context only slightly decreases it. Generally, the impact of adding context is much stronger in the VQ answering than in the reasoning task. Interestingly, contradictory context can sometimes be beneficial, as it helps to minimize the occurrence of silent failures. Additionally, the models continue to show higher attention toward the image than the context, not supporting our prior hypothesis.

\textit{Hypothesis 3 (Including Image Annotations):}
Surprisingly, the image text annotations play a minimal role in enhancing model performance. 
Although the models exhibit increased attention toward annotated images, the positive impact of these annotations on performance metrics and uncertainty reduction is nearly negligible.

We also investigate methods to guide the model to favor one modality over another to observe the effect on VLM performance. While adding redundant textual information can overwhelm the model and decrease accuracy, prompt engineering can improve predictions without, however, strong changes in attention distribution.



\paragraph{Limitations} 
The SI-VQA dataset contains exactly 100 instances. Although relatively small, it has been meticulously handcrafted, with each instance carefully designed to meet key criteria: the question can only be answered using the image; the context provides additional global information either reinforcing the image's consistency or misleading the model; and the annotations are simplified concepts written directly on the image to aid in accurately interpreting the scene. In this work, we also limit our study to seven different modality configurations with specific interventions. For those interested in more specific or advanced interventions, we refer to our ISI Tool, which allows testing of almost all possible forms of interventions. In future research, it would be interesting to replicate this study in the reverse scenario—where text is the primary content required for answering, and the image serves as contextual information—to compare whether similar effects of primary and secondary modalities are observed. We employ semantic entropy as our unique uncertainty measure, considering other measures for free-form text generation, such as token entropy \citep{kadavath2022language,Lindley1956} or self-expressed uncertainty \citep{lin2022teachingmodels,Liao2024AI}, to be significantly less suitable.
They either only consider local, token-level uncertainty or rely on the model's potentially biased self-assessment, neither adequately reflecting the overall semantic uncertainty. 
We validated the expected behavior of semantic entropy across the different configurations using the AUGRC metric in \autoref{apx:val_se}.

\paragraph{Conclusion}
This study is the first to systematically examine the role of contextual information in VQA, evaluating the results based on diverse metrics and distinguishing between answering and reasoning tasks. 
Leveraging the well-curated SI-VQA dataset and the ISI tool—our interactive, ready-to-use interface—our work aims to provide a deeper understanding of VLM behavior and the influence of each modality.
Moreover, we show that our results can also be used to understand and detect model failure in free-form text generation, and setting the stage for future analyses of modality integration across various VLM tasks and the development of VL datasets tailored for this objective.

\newpage
\bibliography{iclr2025_conference}
\bibliographystyle{iclr2025_conference}

\newpage

\appendix
\section{Modality Attention Aggregation}\label{apx:attention}

From the VLMs, we extract an attention matrix $A_t\in\mathbb{R}^{t\times t}$ for each output token $t$, where the matrix size grows with the number of predicted tokens. Each row represents a token as a query, and each column corresponds to a token as a key. Therefore, each row $r$ contains the attention coefficients of each token $i$ with respect to the token $r$: $A[r,i] = \alpha_{r,i} = q_r^{\top} k_i$. For each token $t$, we focus only on the attention of preceding tokens, represented by $v_t^{\top}=A[-1]\in\mathbb{R}^{1\cdot t}$. $v_t$ is normalized so that the attention coefficients of all preceding tokens to $t$ sum to 1.

This process is repeated for all output tokens $t\in [1,T]$ where $T$ is the output length, yielding all normalized attention vectors $v_t, t\in [1,T]$. These are averaged to produce the final attention vector for the VLM output—$V_A\in\mathbb{R}^{N_0}$ for answers and $V_R\in\mathbb{R}^{N_0+T_A+N_1}$ for reasoning. Here, $N_0$ is the size of the input prompt including image tokens, question, and eventually context, $T_A$ is the answer length, and $N_1$ is the size of the second prompt asking for an explanation.

To calculate the attention given to different modalities, we sum the attention coefficients based on the token positions in the averaged attention vector $V_A$ or $V_R$, resulting in normalized relevance scores: $R_I$ for the image, $R_Q$ for the question, and $R_C$ for the context, which sum to 1, i.e., $R_I+R_Q+R_C=1$. This is done by adding hooks into the LLaVA architecture to capture the start and end positions of the question, image, and context tokens. Due to dynamic high resolution, the number of image tokens can vary significantly even with minimal image perturbations.

\begin{algorithm}
\caption{Attention Attribution Computation}
\begin{algorithmic}[1]

\State \textbf{Input:} Attention matrix $A_t \in \mathbb{R}^{t \times t}$ for each token $t$, Token positions for Image ($I$), Question ($Q$), and Context ($C$), Length of output $T$
\State \textbf{Output:} Normalized relevance scores $R_I, R_Q, R_C$

\For{$t \in [1, T]$} 
    \State Extract attention for preceding tokens: $v_t^{\top} = A[-1] \in \mathbb{R}^{1 \times t}$
    \State Normalize attention coefficients: $v_t \gets \frac{v_t}{\sum_{i=1}^{t} v_t[i]}$
\EndFor

\State Compute average attention vector: $V = \frac{1}{T} \sum_{t=1}^{T} v_t^T$

\State Compute total attention for each modality:
\State $R_I \gets \sum_{i \in I} V[i]$
\State $R_Q \gets \sum_{i \in Q} V[i]$
\State $R_C \gets \sum_{i \in C} V[i]$

\State Normalize relevance scores:
\State $R_I \gets \frac{R_I}{R_I + R_Q + R_C}$
\State $R_Q \gets \frac{R_Q}{R_I + R_Q + R_C}$
\State $R_C \gets \frac{R_C}{R_I + R_Q + R_C}$

\State \textbf{Return:} $R_I, R_Q, R_C$

\end{algorithmic}
\end{algorithm}

\section{Hyperparameter and Evaluation Prompts} \label{apx:hyperparameters}
In reasoning quality evalution, GPT-4o is prompted to rate the reasoning from 0 to 10, using the prompt: ``Rate the explanation's quality from 0 to 10. Give 10 for detailed, well-argued, and correct explanations. Give 0 for a poorly reasoned, wrong, or single-word explanation based on the question and image. Don't rate too harshly, use the full scale and output only the final score''.
 During uncertainty computation, the number of the sampled outputs and the sampling temperaturee $T$ are set to 10 and 0.9 respectivly. We use conditional probabilistic sampling.
 See \autoref{apx:attention} for the attention attribution implementation.

 In the prompt engineering ablation study in \autoref{sec:rebalancing} we use the prompt ``Answer the question only with Yes or No. Answer based on the context provided in the text.'' for the answer, followed by ``Explain your answer based on the context provided in the text:'' for reasoning.
 
\section{SI-VQA Dataset Baseline} \label{apx:vqa_baseline}

In the baseline configuration of the SI-VQA Dataset, the image tokens carry no meaningful information. We experiment with several methods to remove image data: (1) replacing the image with black pixels, (2) adding random Gaussian noise centered on (0,0,0), or (3) omitting the image tokens entirely. In (1) and (2), the image size is by default set to $256\times256$ pixels. Across our experiments, we observed similar results when using black pixels or noise. However, when no image input is provided, the model produces incoherent predictions and reasoning, often generating ``$\backslash$n''. Scenario (3) is out-of-distribution for any VLMs, leading to nonsensical outputs, as the model was trained exclusively in a multimodal context and has never encountered unimodal (text-only) input.
\vspace{1cm}
\begin{tcolorbox}[colback=black!5!white, title=(1) Baseline configuration with black image]
\scriptsize
\linespread{1.3}\selectfont

\begin{wrapfigure}{r}{0.07\textwidth}
    \centering
    \vspace{-1.5\intextsep}%
    \includegraphics[width=0.07\textwidth]{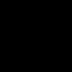}
\end{wrapfigure}
\vspace{.8em}\textbf{Question:} Is there a yellow and black frog in this school book?\\
\textbf{Ground truth answer:} Yes\\
\rule{\linewidth}{0.2mm} 
\textbf{Answer:} No\\
\textbf{Reasoning:} The image you provided is completely black, which does not allow for the identification of any objects or creatures, including a yellow and black frog.
\end{tcolorbox}

\begin{tcolorbox}[colback=black!5!white, title=(2) Baseline configuration with noisy image]
\scriptsize
\linespread{1.3}\selectfont
\begin{wrapfigure}{r}{0.07\textwidth}
    \centering
    \vspace{-1.5\intextsep}%
    \includegraphics[width=0.07\textwidth]{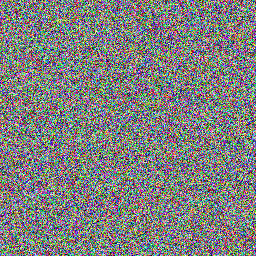}
\end{wrapfigure}
\vspace{.8em}\textbf{Question:} Is there a yellow and black frog in this school book?\\
\textbf{Ground truth answer:} Yes\\
\rule{\linewidth}{0.2mm} 
\textbf{Answer:} No\\
\textbf{Reasoning:} The image is a close-up of a colorful, pixelated background. There is no frog visible in the image.
\end{tcolorbox}

\begin{tcolorbox}[colback=black!5!white, title=(3) Baseline configuration with no image token]
\scriptsize
\linespread{1.3}\selectfont
\vspace{.8em}\textbf{Question:} Is there a yellow and black frog in this school book?\\
\textbf{Ground truth answer:} Yes\\
\rule{\linewidth}{0.2mm} 
\textbf{Answer:} 是 이  이 $\backslash$n           $\backslash$n Is a $\backslash$n$\backslash$n \\
\textbf{Reasoning:} \hspace{1cm} 10
\end{tcolorbox}

\newpage

\section{SI-VQA Benchmark Examples}

Each instance of the SI-VQA Dataset was double-checked and validated by three individuals.

\begin{tcolorbox}[colback=black!5!white, title=Boiling.jpg]
\scriptsize
\linespread{1.3}\selectfont

\textbf{Question:} Is the water hot?\\
\textbf{Ground truth answer:} Yes\\
\textbf{\textcolor[HTML]{3366CC}{LLaVA 7B:}} No. The water is not hot.

\textbf{\textcolor[HTML]{0099C5}{LLaVA-Next 7B:}} No. The image you provided is too dark and lacks any discernible features to determine if the water is hot or not.

\vspace{.2em}\rule{\linewidth}{0.2mm}\vspace{.2em}
\begin{wrapfigure}{r}{0.07\textwidth}
    \centering
    \vspace{-1.2\intextsep}%
    \includegraphics[width=0.07\textwidth]{figures/boiling.jpg}
\end{wrapfigure}
\textbf{Question:} Is the water hot?

\textbf{\textcolor[HTML]{3366CC}{LLaVA 7B:}} No. The water in the pitcher is cold.

\textbf{\textcolor[HTML]{0099C5}{LLaVA-Next 7B:}} No. The image shows a pitcher of water with ice cubes, which suggests that the water is cold.

\rule{\linewidth}{0.2mm} 
\begin{wrapfigure}{r}{0.07\textwidth}
    \centering
    \vspace{-1.8\intextsep}%
    \includegraphics[width=0.07\textwidth]{figures/boiling.jpg}
\end{wrapfigure}
\textbf{Question:} Is the water hot?

\textbf{Complementary Context:} Kettles, essential for boiling water, are commonly used for making tea, coffee, or instant meals, and come in various designs, including traditional stove top and modern electric models.

\textbf{\textcolor[HTML]{3366CC}{LLaVA 7B:}} Yes. The water in the kettle is hot.

\textbf{\textcolor[HTML]{0099C5}{LLaVA-Next 7B:}} No. The image shows a water kettle with water inside it, but it is not hot. The water appears to be at room temperature or slightly cooler, as indicated by the condensation on the outside of the kettle.

\rule{\linewidth}{0.2mm}
\begin{wrapfigure}{r}{0.07\textwidth}
    \centering
    \vspace{-1\intextsep}%
    \includegraphics[width=0.07\textwidth]{figures/boiling.jpg}
\end{wrapfigure}
\textbf{Question:} Is the water hot?

\textbf{Contradictory Context:} Glass pitchers are often used to serve ice-cold beer, making them a popular choice in bars and restaurants for sharing drinks in a casual, communal setting.

\textbf{\textcolor[HTML]{3366CC}{LLaVA 7B:}} No. The water in the pitcher is not hot. It is ice-cold, which is a common feature of glass pitchers used for serving beer.

\textbf{\textcolor[HTML]{0099C5}{LLaVA-Next 7B:}} No. The image shows a glass pitcher filled with ice and water, which is not hot. The ice is used to keep the water cold, and the pitcher is designed to serve multiple people at once. This type of pitcher is commonly used for serving cold beverages, such as water or iced tea, in a social setting.

\rule{\linewidth}{0.2mm}
\begin{wrapfigure}{r}{0.07\textwidth}
    \centering
    \vspace{-1\intextsep}%
    \includegraphics[width=0.07\textwidth]{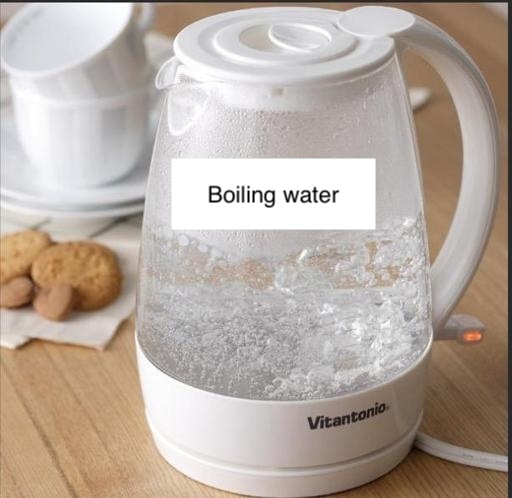}
\end{wrapfigure}
\textbf{Question:} Is the water hot?

\textbf{\textcolor[HTML]{3366CC}{LLaVA 7B:}} No. Boiling water is water that is heated to the point of boiling, which is the temperature at which water vaporizes. In this image, the water in the kettle is not hot, as it is described as "boiling water" and "cold water."

\textbf{\textcolor[HTML]{0099C5}{LLaVA-Next 7B:}} Yes. The image shows a water pitcher with the label "Boiling water" on it, which suggests that the water inside is being heated to boiling point. The presence of steam and the fact that the water is being heated in a pitcher designed for boiling water further supports the idea that the water is hot.

\rule{\linewidth}{0.2mm}
\begin{wrapfigure}{r}{0.07\textwidth}
    \centering
    \vspace{-1\intextsep}%
    \includegraphics[width=0.07\textwidth]{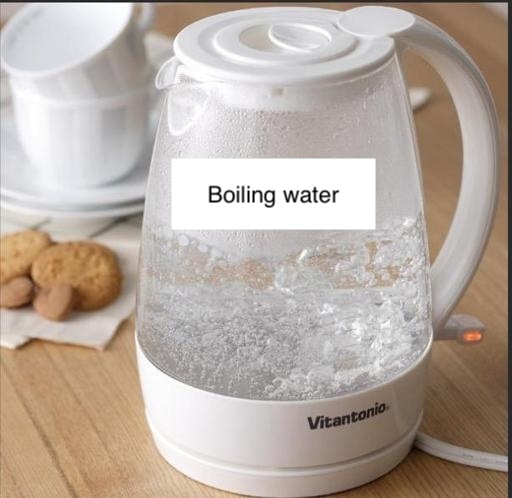}
\end{wrapfigure}
\textbf{Question:} Is the water hot?

\textbf{Complementary Context:} Kettles, essential for boiling water, are commonly used for making tea, coffee, or instant meals, and come in various designs, including traditional stove top and modern electric models.

\textbf{\textcolor[HTML]{3366CC}{LLaVA 7B:}} Yes. The water in the kettle is hot, as indicated by the steam rising from it.

\textbf{\textcolor[HTML]{0099C5}{LLaVA-Next 7B:}} No. The image shows a water kettle with a label that reads "Boiling water." However, the water in the kettle appears to be cold, as indicated by the condensation on the outside of the kettle. This suggests that the water has not yet reached its boiling point.

\rule{\linewidth}{0.2mm}
\begin{wrapfigure}{r}{0.07\textwidth}
    \centering
    \vspace{-1\intextsep}%
    \includegraphics[width=0.07\textwidth]{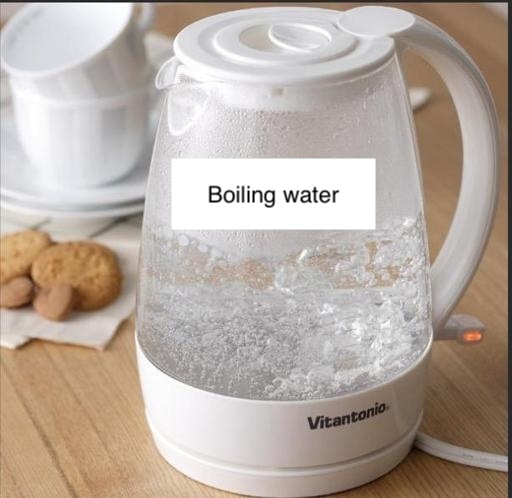}
\end{wrapfigure}
\textbf{Question:} Is the water hot?

\textbf{Contradictory Context:} Glass pitchers are often used to serve ice-cold beer, making them a popular choice in bars and restaurants for sharing drinks in a casual, communal setting.

\textbf{\textcolor[HTML]{3366CC}{LLaVA 7B:}} No. The water in the pitcher is not hot. It is ice-cold, which is a common feature of glass pitchers used for serving beer.

\textbf{\textcolor[HTML]{0099C5}{LLaVA-Next 7B:}} No. The image shows a water pitcher with the label "Boiling water" on it, but the water inside appears to be cold. The context of the image, which includes a wooden table and a few cookies, suggests that the water is being used for a hot beverage, such as tea or coffee. However, the water itself is not boiling at the moment the image was taken.
\end{tcolorbox}

\newpage

\begin{tcolorbox}[colback=black!5!white, title=Painting\_asia.jpg]
\scriptsize
\linespread{1.3}\selectfont

\textbf{Question:} Is the painting of european descent?\\
\textbf{Ground truth answer:} No\\
\textbf{\textcolor[HTML]{3366CC}{LLaVA 7B:}} No. The painting is of African descent.

\textbf{\textcolor[HTML]{0099C5}{LLaVA-Next 7B:}} No. The image you provided is too dark and lacks detail to determine if it is a painting of European descent. The image is not clear enough to make an accurate assessment.

\vspace{.2em}\rule{\linewidth}{0.2mm}\vspace{.2em}
\begin{wrapfigure}{r}{0.07\textwidth}
    \centering
    \vspace{-1.2\intextsep}%
    \includegraphics[width=0.07\textwidth]{figures/painting_asia.jpg}
\end{wrapfigure}
\textbf{Question:} Is the painting of european descent?

\textbf{\textcolor[HTML]{3366CC}{LLaVA 7B:}} No. The painting is of Asian descent.

\textbf{\textcolor[HTML]{0099C5}{LLaVA-Next 7B:}} No. The painting is not of European descent. It appears to be a traditional East Asian artwork, likely depicting a scene from a historical period in China or Japan, given the style of clothing and the architectural elements.

\rule{\linewidth}{0.2mm} 
\begin{wrapfigure}{r}{0.07\textwidth}
    \centering
    \vspace{-1.8\intextsep}%
    \includegraphics[width=0.07\textwidth]{figures/painting_asia.jpg}
\end{wrapfigure}
\textbf{Question:} Is the painting of european descent?

\textbf{Complementary Context:} Scholars and students engaged in study and teaching within an ancient Chinese educational setting. The intricate details of the architecture and the serene outdoor environment emphasize the importance of learning and intellectual pursuit in classical Chinese culture.

\textbf{\textcolor[HTML]{3366CC}{LLaVA 7B:}} Yes. The painting is of an ancient Chinese educational setting, featuring scholars and students engaged in study and teaching. The intricate details of the architecture and the serene outdoor environment emphasize the importance of learning and intellectual pursuit in classical Chinese culture.

\textbf{\textcolor[HTML]{0099C5}{LLaVA-Next 7B:}} No. The painting is not of European descent. It depicts a traditional Chinese scene, with scholars and students engaged in study and teaching within an ancient Chinese educational setting. The architecture and the serene outdoor environment are characteristic of Chinese art and culture, and the painting is not of European origin.

\rule{\linewidth}{0.2mm}
\begin{wrapfigure}{r}{0.07\textwidth}
    \centering
    \vspace{-1\intextsep}%
    \includegraphics[width=0.07\textwidth]{figures/painting_asia.jpg}
\end{wrapfigure}
\textbf{Question:} Is the painting of european descent?

\textbf{Contradictory Context:} In an old German schoolhouse, students diligently engage in their lessons, surrounded by the timeless charm of wooden desks, slate chalkboards, and the warm glow of afternoon sunlight filtering through the tall, arched windows.

\textbf{\textcolor[HTML]{3366CC}{LLaVA 7B:}} Yes. Yes.

\textbf{\textcolor[HTML]{0099C5}{LLaVA-Next 7B:}} Yes. The painting depicts a scene that is characteristic of traditional European schools, with students seated at wooden desks, using slate chalkboards, and the presence of tall, arched windows suggests a European architectural style. The clothing of the students and the overall composition of the painting are consistent with European art and culture.

\rule{\linewidth}{0.2mm}
\begin{wrapfigure}{r}{0.07\textwidth}
    \centering
    \vspace{-1\intextsep}%
    \includegraphics[width=0.07\textwidth]{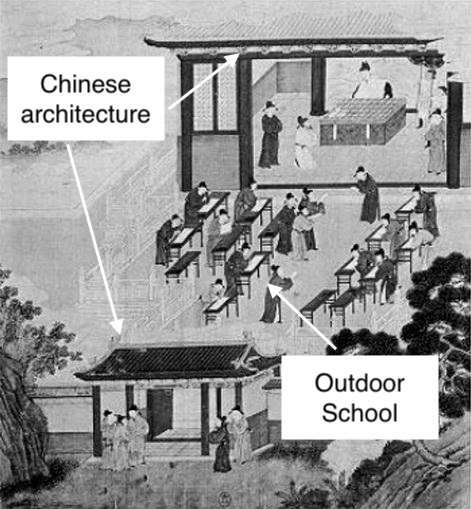}
\end{wrapfigure}
\textbf{Question:} Is the painting of european descent?

\textbf{\textcolor[HTML]{3366CC}{LLaVA 7B:}} No. The painting is of Chinese descent.

\textbf{\textcolor[HTML]{0099C5}{LLaVA-Next 7B:}} No. The painting is not of European descent. It depicts a traditional Chinese scene, with Chinese architecture, clothing, and people engaged in activities such as writing and reading. The style of the painting, including the clothing and the architectural elements, is indicative of a Chinese setting, which suggests that the painting is not of European descent.

\rule{\linewidth}{0.2mm}
\begin{wrapfigure}{r}{0.07\textwidth}
    \centering
    \vspace{-1\intextsep}%
    \includegraphics[width=0.07\textwidth]{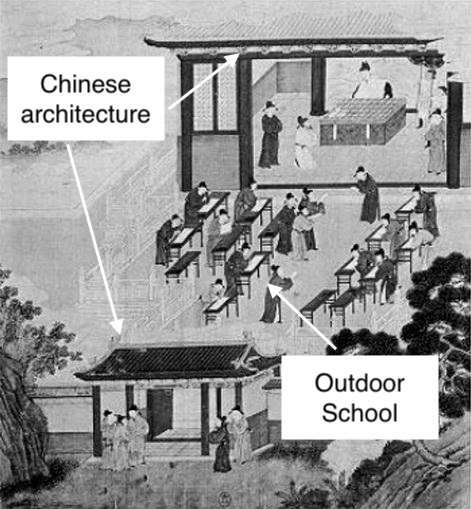}
\end{wrapfigure}
\textbf{Question:} Is the painting of european descent?

\textbf{Complementary Context:} Scholars and students engaged in study and teaching within an ancient Chinese educational setting. The intricate details of the architecture and the serene outdoor environment emphasize the importance of learning and intellectual pursuit in classical Chinese culture.

\textbf{\textcolor[HTML]{3366CC}{LLaVA 7B:}} No. The painting is of Chinese descent, depicting a scene of scholars and students engaged in study and teaching within an ancient Chinese educational setting. The intricate details of the architecture and the serene outdoor environment emphasize the importance of learning and intellectual pursuit in classical Chinese culture.

\textbf{\textcolor[HTML]{0099C5}{LLaVA-Next 7B:}} No. The painting is not of European descent. It depicts a traditional Chinese scene, with Chinese architecture, clothing, and hairstyles, which are characteristic of East Asian art. The style of the painting, including the brushwork and the subject matter, are indicative of Chinese art and culture.

\rule{\linewidth}{0.2mm}
\begin{wrapfigure}{r}{0.07\textwidth}
    \centering
    \vspace{-1\intextsep}%
    \includegraphics[width=0.07\textwidth]{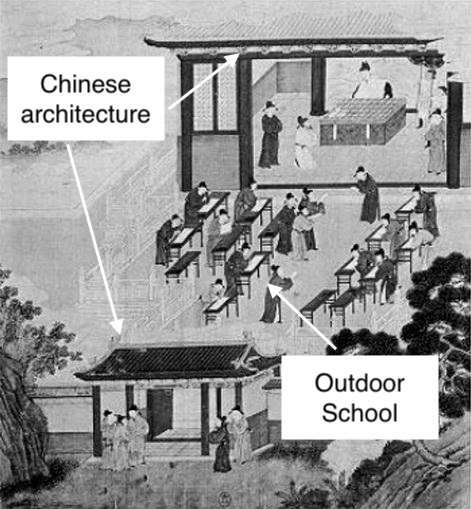}
\end{wrapfigure}
\textbf{Question:} Is the painting of european descent?

\textbf{Contradictory Context:} In an old German schoolhouse, students diligently engage in their lessons, surrounded by the timeless charm of wooden desks, slate chalkboards, and the warm glow of afternoon sunlight filtering through the tall, arched windows.

\textbf{\textcolor[HTML]{3366CC}{LLaVA 7B:}} No. The painting is of Chinese descent, depicting a scene from a Chinese school.

\textbf{\textcolor[HTML]{0099C5}{LLaVA-Next 7B:}} Yes. The painting depicts a traditional European schoolhouse, with students seated at wooden desks and chalkboards, and the architecture and style of the building suggest a European influence. The presence of tall, arched windows and the overall composition of the scene are characteristic of European art and architecture.
\end{tcolorbox}

\section{ISI Tool}\label{apx:isitool}

\begin{figure*}[!ht]
    \centering
    \includegraphics[trim=0.0cm 20cm 0.0cm 0cm,clip, width=\linewidth, page = 1]{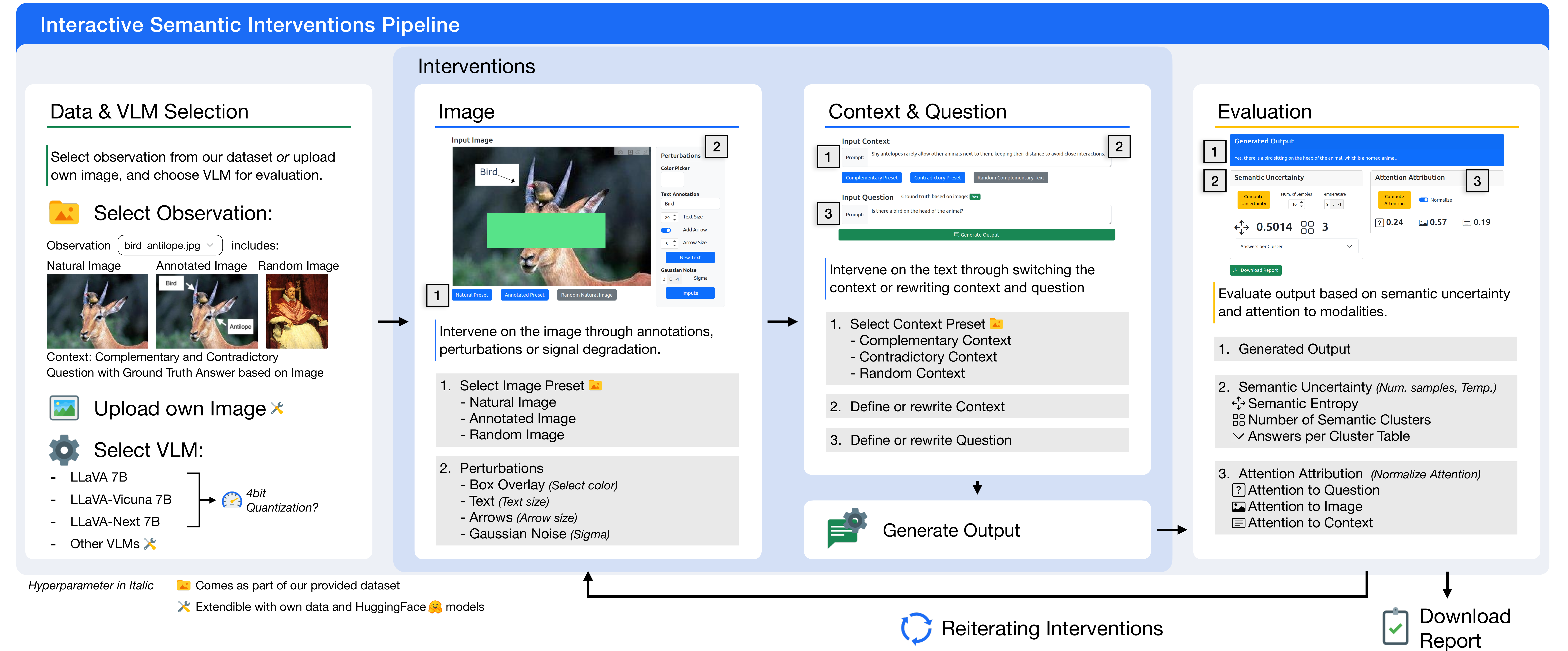}
    \caption{Illustration of the evaluation pipeline used in the ISI for VLMs to enable interactive exploration of VLM behavior under various scenarios. It consists of three main stages: 1) Data \& VLM Selection: Users choose an observation either from the provided SI-VQA Dataset or upload their own, and select a VLM for evaluation. 2) Interventions on Image, Context \& Question: The selected image can be altered through presets or perturbations, and the context or question can be edited or also switched with presets. 3) Evaluation: The output is analyzed for semantic uncertainty and attention attribution, allowing for iterative refinement of interventions.} 
    \label{fig:main-tool}
\end{figure*}

The interactive tool can be used to analyze VLMs with the provided SI-VQA Dataset and follows a main pipeline that consists of three main steps: 1) Data \& Model Selection 2) Interventions on Image, Context, and Question, and 3) Evaluation. \autoref{fig:main-tool} gives an overview of this pipeline. 

\subsection{General Information}\label{ss:general-info}
\paragraph{Users}
The application is catered toward researchers, developers, and other users with a basic understanding of VLMs, who are interested in interpreting model behavior through semantic interventions on VLMs. By enabling fast-paced iterations in a human-in-the-loop scenario, it allows the building of intuitions before scaling experiments in large-scale projects.

\paragraph{System Requirements} 
ISI for VLMs is an interactive tool embedded in a locally hosted web application requiring a computer with sufficient VRAM for VLM inference. 
The minimum required VRAM for a 4bit-quantized LLAVA 7B model is around 8GB while LLaVA-Vicuna and LLaVA-Next require 12GB. The computation of the semantic entropy with the DeBERTa model requires an additional 7GB. The exact amount of VRAM depends on the amount of input tokens.

\subsection{Interactive Semantic Intervention Pipeline}\label{ss:pipeline}

\paragraph{Data \& Model Selection:}
As the first step, a user either chooses an observation from the SI-VQA Dataset or uploads their custom image. Each observation from the dataset includes an image, corresponding context, and a question with a ground truth answer, as well as the presets for the annotated images and contradictory and complementary context. The corresponding image, context, and question are displayed.
In the next step, the user selects a VLM (LLaVA, LLaVA-Vicuna, LLaVA-Next) and the number of parameters (7B, 11B, 32B) in two separate drop-down menus. 4-bit quantization can be enabled to reduce the computational load and VRAM requirements. 

\paragraph{Interventions on the Image:}
For interventions on the image, ISI allows the user two main functionalities.
First, on the proposed SI-VQA Dataset the user can select for each observation three different image presets (natural image without modifications, annotated image with hand-crafted annotations, and random natural image from the dataset) by selecting the respective buttons.
Second, ISI allows perturbing the image directly in the tool by overlaying boxes with selectable colors, inserting and modifying text, adding directional arrows, and introducing Gaussian noise with adjustable noise values. 

\paragraph{Interventions on Context \& Question:} To facilitate the user’s ability to observe how various contexts and questions affect the model’s performance two functionalities are supported.
In the proposed SI-VQA Dataset, users can choose from three distinct context presets—complementary, contradictory, or random—by selecting the respective button, automatically updating the content in the text input fields.
Additionally, the user can manually edit the context and question in these fields.

\paragraph{Evaluation}
The evaluation is designed to enable quantitative analysis of how interventions on image and text impact the behavior of the selected VLM. At the top, the current input is visualized to always relate the evaluation results to the correct input. Below,  the generated output, semantic uncertainty, and attention attribution are shown. For computational reasons, each evaluation can be started separately.

The semantic uncertainty tab allows users to evaluate model uncertainty by clustering sampled outputs according to their semantic meaning. This feature highlights the range of semantic differences in the outputs and calculates semantic entropy, providing a comprehensive view of the model's overall uncertainty. It displays key metrics, such as semantic entropy and the number of semantic clusters, which are influenced by adaptable hyperparameters like the number of samples and the sampling temperature. For deeper exploration, the "Answers per Cluster" dropdown provides a table displaying all sampled answers along with their assigned semantic clusters. This table enables users to examine the full range of generated outputs and understand the semantic similarities within each cluster. To evaluate the significance of each of the three inputs during generation, the attention attribution tab displays the absolute or relative attention assigned to the question, context, and image input tokens.

To provide a contextual understanding of the current observation, the tool additionally displays average values for attention attribution and semantic entropy across the entire SI-VQA Dataset based on each VLM architecture. These averages are shown when hovering over the relevant values.

\paragraph{Export}
The results of one iteration can be exported as a PDF to facilitate the systematic collection of example cases for further analysis and to support the transition from initial qualitative insights to small-scale quantitative evaluation. After the analysis, users can download a comprehensive report that includes the image, context, question, detailed model setup, hyperparameters, and all computed evaluation metrics.

\newpage
\section{4Bit Quantization} \label{apx:4bit}

\begin{figure}[H]
    \centering
    \includegraphics[page=9, trim=0cm 50cm 20cm  0cm, clip, width=\linewidth]{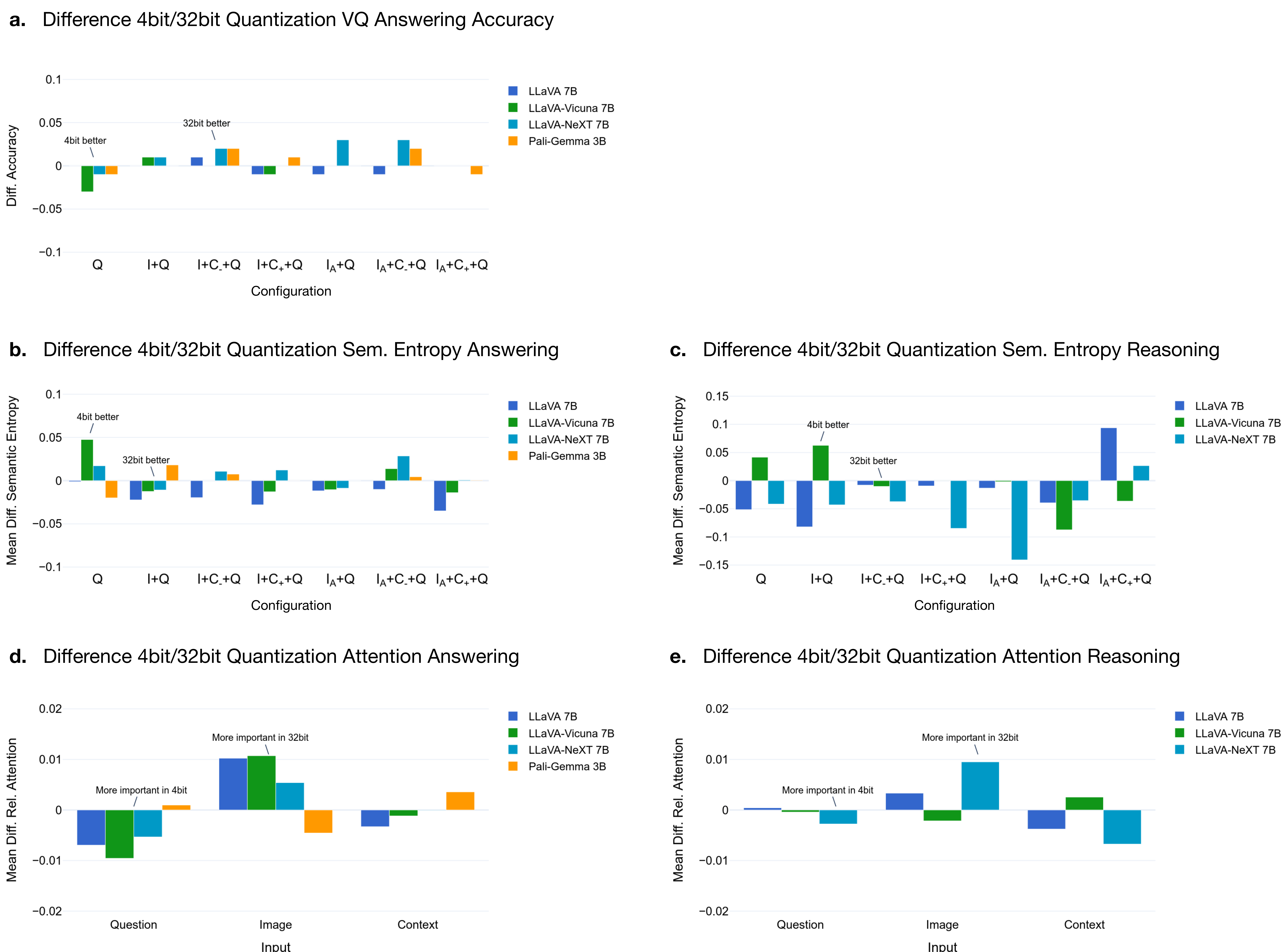}
     \caption{Difference in performance between the 32Bit and 4Bit quantized versions of the models for a. VQA accuracy, b. VQA semantic entropy, c. reasoning semantic entropy, d. VQA attention attribution, and e. reasoning attention attribution.}
    \label{fig:4bit}
\end{figure}

To quantify the effect of different model sizes, we additionally perform all experiments also with the same models but 4Bit quantized, as there are no, e.g., 3B parameter LLaVA versions. \autoref{fig:4bit} shows the difference in results for all five experiments between the 32Bit and 4Bit models. To our surprise, the difference in accuracy is not that large. Results for the question-only configuration are not meaningful as both models randomly guess. However, in terms of model uncertainty, the 32Bit model usually scores better. Mean differences in attention distribution are almost neglectable as they are at a maximum of 0.01 percentage points. The results show that for simple VQA, quantized models can achieve almost similar accuracy than their significantly larger 32Bit counterparts. 

\section{Validating Semantic Entropy} \label{apx:val_se}

Any assumptions made on the uncertainty of the models should be reflected in the AUGRC.
We observe in \autoref{fig:failure} that in the case of contradicting image and context (Q+I+C$_-$) the AUGRC goes down for all models, reducing the overconfidence in wrong classified samples, which is correctly captured by the semantic entropy.
As the model makes more mistakes in configuration Q+I+C$_-$, the set size of wrong classified samples is larger. 
In the case of Q+I or Q+I+C$_+$ the AUGRC rises again as there is no confidence reducing context.
Additionally, the accuracy is higher in the case of Q+I+C$_+$, reducing the set size of wrong classified samples.
This empirical evaluation shows we can use the AUGRC to quantify and evaluate the performance of semantic entropy for model failure detection. 
To our knowledge, this is the first time semantic entropy has been evaluated for failure detection.

\section{Additional Results}

\subsection{VQA Accuracy and Quality} \label{apx:vaqacc_results}

\begin{figure}[H]
    \centering
    \includegraphics[page=5, trim=0cm 5cm 0cm  0cm, clip, width=\linewidth]{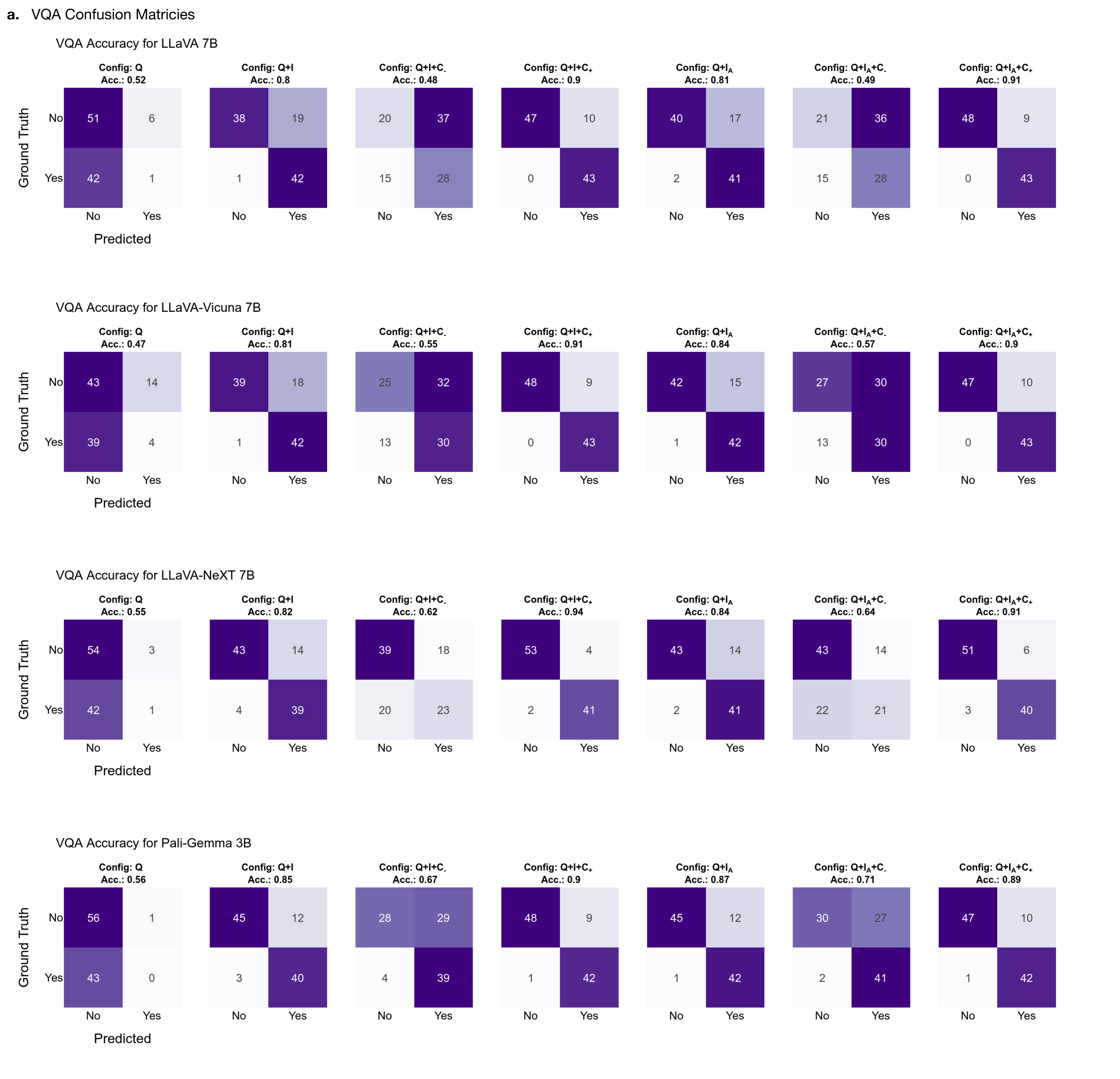}
     \caption{Confusion matrices and accuracy values for all model architectures and configurations.}
    \label{fig:vqa_acc_conf}
\end{figure}

\begin{figure}[ht!]
    \centering
    \includegraphics[page=6, trim=0cm 95cm 77cm  4cm, clip, width=.7\linewidth]{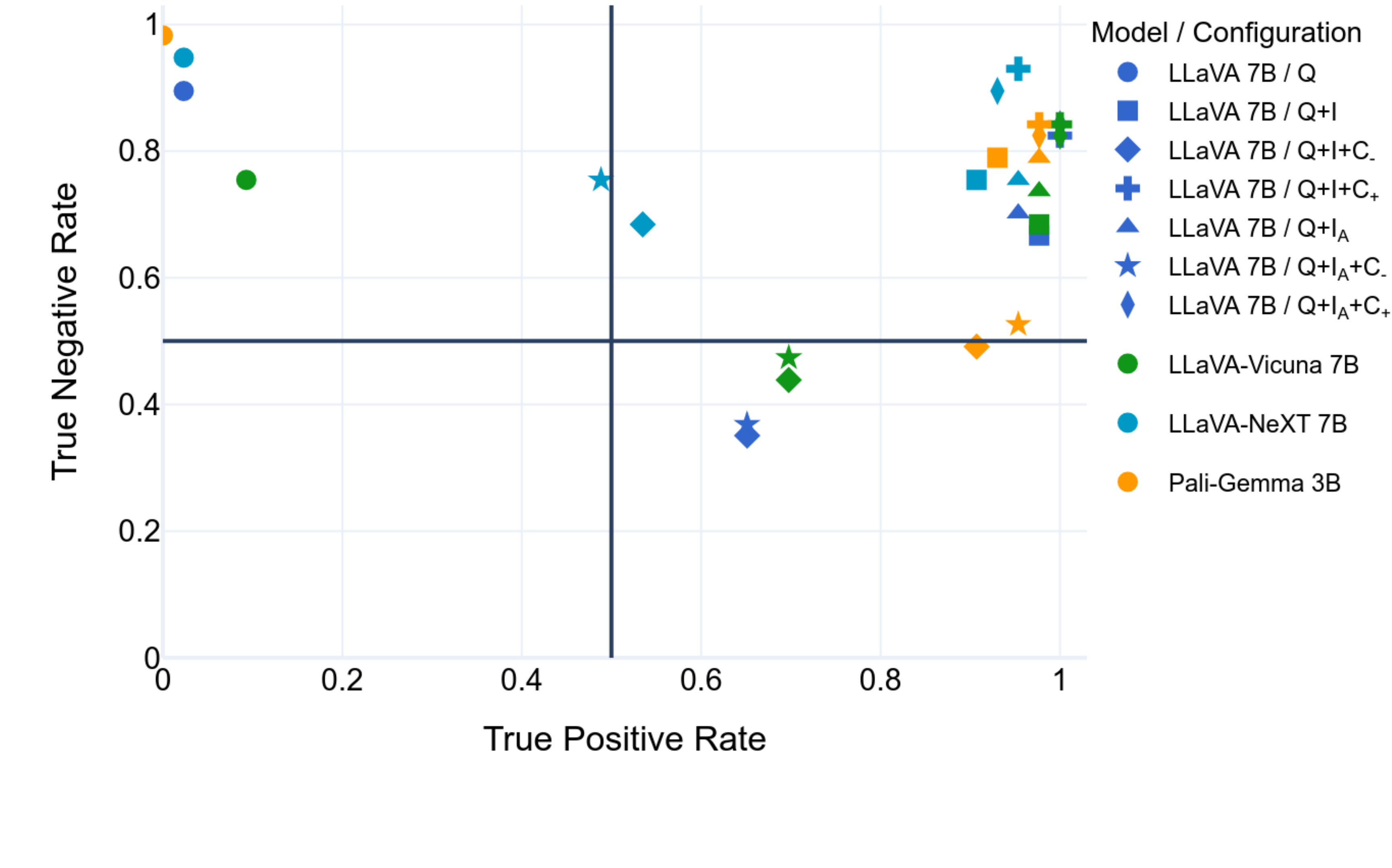}
     \caption{True Negative Rate versus the True Positive Rate in the VQA task of all model architectures and configurations. Values below 0.5 on one axis indicate worse than random guessing performance.}
\end{figure}

\begin{figure}[ht!]
    \centering
    \includegraphics[page=6, trim=65cm 95cm 5cm  4cm, clip, width=.7\linewidth]{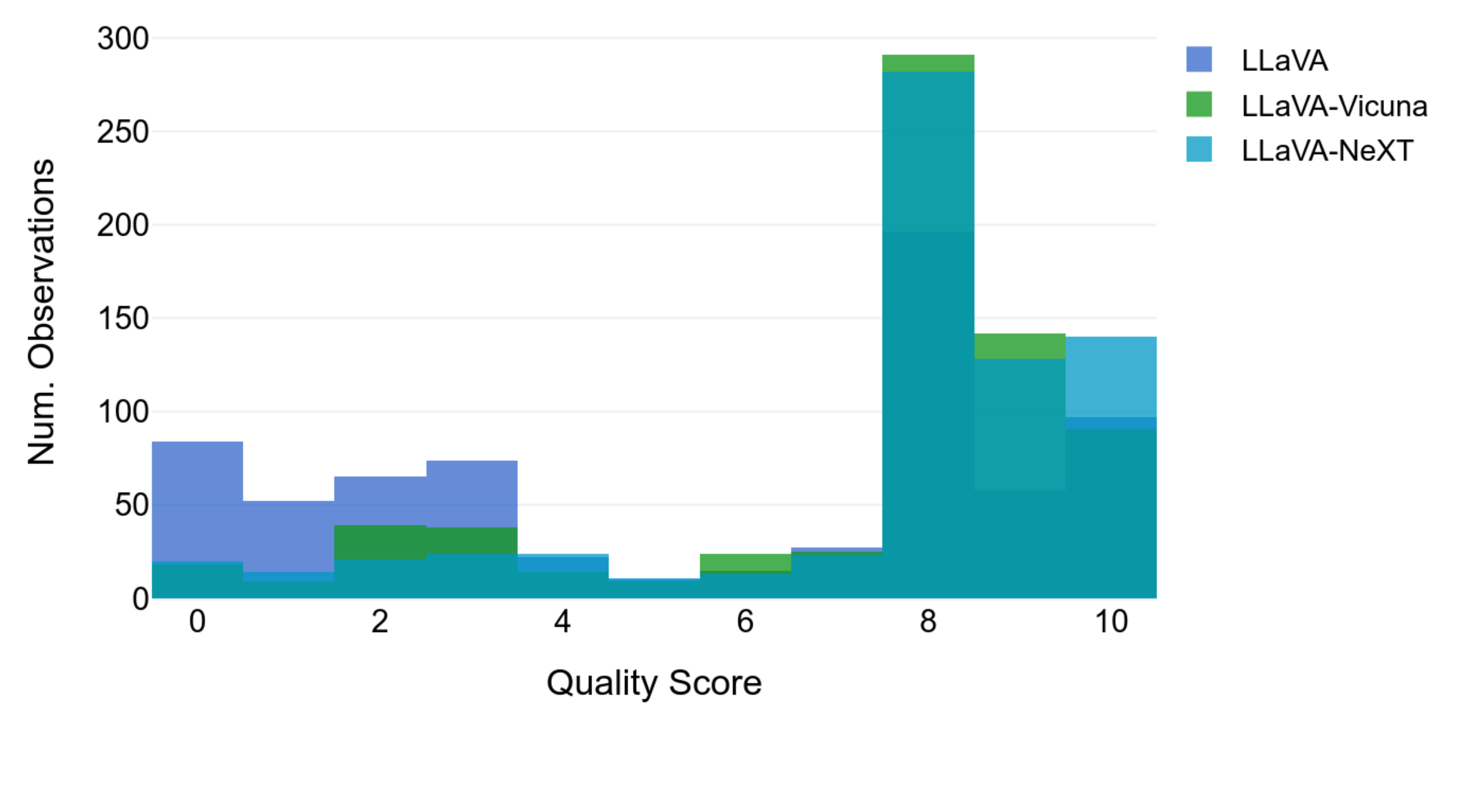}
     \caption{Distribution of reasoning quality scores by GPT-4o for all LLaVA architectures. We observe lower values for the standard LLaVA and a high bias towards the quality score of ``8'' for all models.}
\end{figure}

\subsection{Semantic Entropy} \label{apx:sem_entr_results}

\begin{figure}[H]
    \centering
    \includegraphics[page=7, trim=0cm 15cm 40cm  0cm, clip, width=\linewidth]{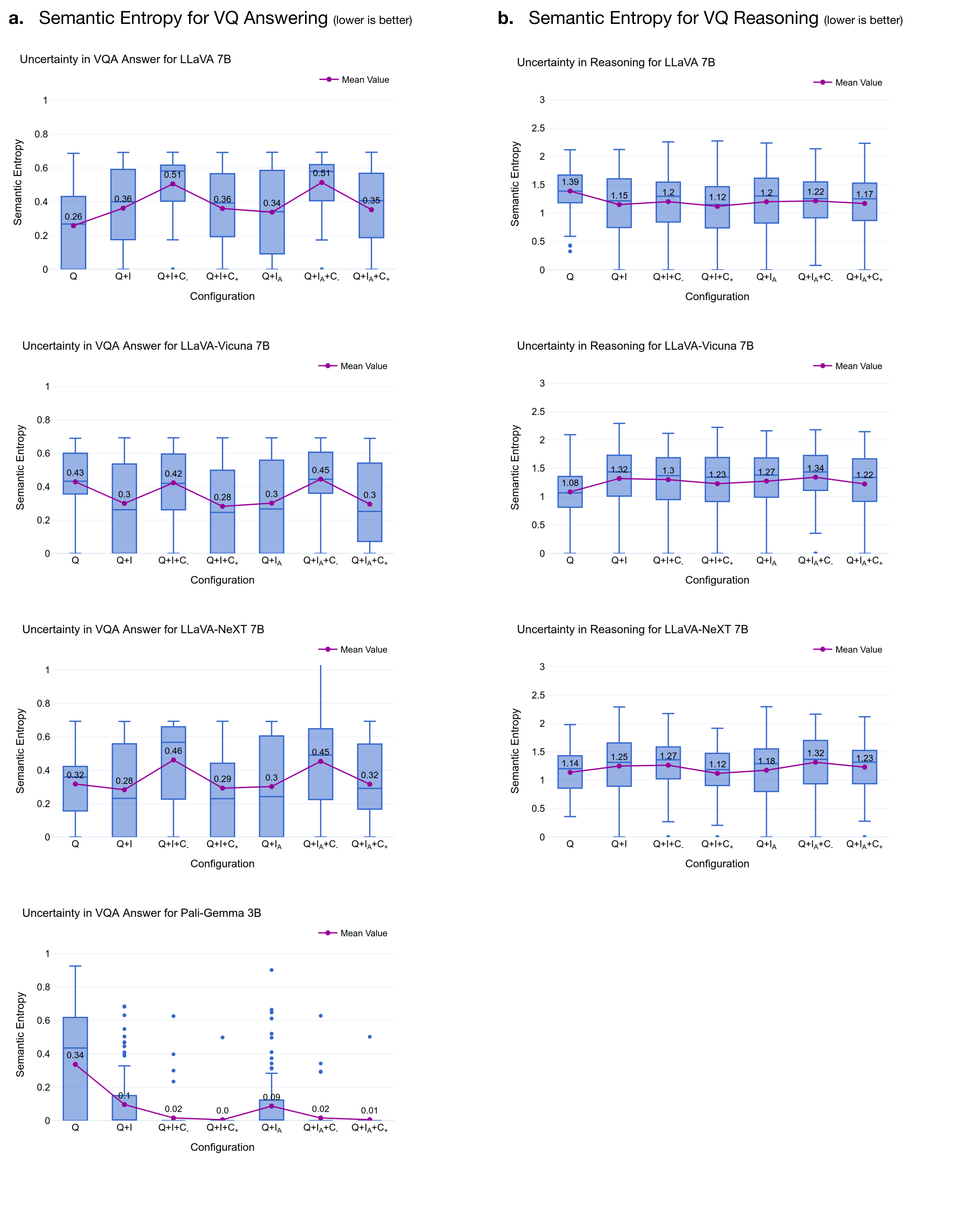}
    \vspace{-4em}
     \caption{Distribution of semantic entropy for VQ answering and reasoning task for all model architectures and configurations.}
    \label{fig:sem_entr}
\end{figure}

\subsection{Attention} \label{apx:attention_results}

\begin{figure}[H]
    \centering
    \includegraphics[page=8, trim=0cm 55cm 10cm  0cm, clip, width=\linewidth]{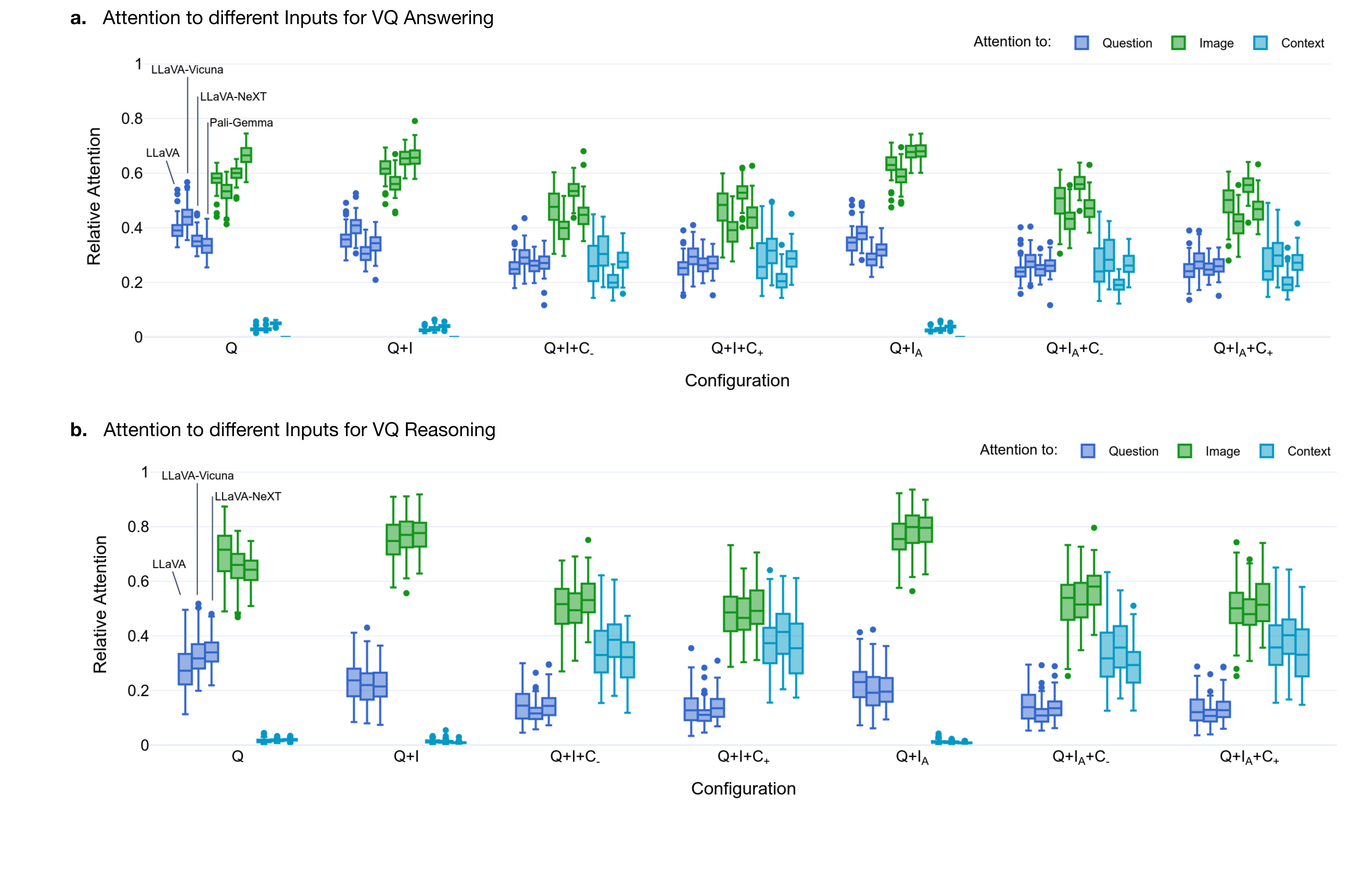}
     \caption{Attention attribution to the question (Q), image (I), and context (C$_{+/-}$) by the three LLaVA models and PaliGemma and the seven modality configurations, during the answering (a.) and the reasoning (b.) in the VQA tasks. Variance across samples is greater for the reasoning than the answering process.}
    \label{fig:atten_var}
\end{figure}

\subsection{Pearson Correlation between Attention and Accuracy} \label{apx:acc_corr}

\begin{table}[H]
    \centering
    \caption{The Pearson correlation coefficients (PCC) between the attention to the different inputs and the accuracy per sample. Correlation is between -0.07 and 0.9 for all inputs and model architectures and can fluctuate between architectures.}
    \setlength{\tabcolsep}{12pt} 
    \renewcommand{\arraystretch}{2} 
    \begin{tabular}{lcccc}
        \toprule
        \textbf{Model Architecture:} & \textbf{LLAVA 7B} & \textbf{LLAVA-Vicuna 7B} & \textbf{LLAVA-NeXT 7B} & \textbf{PaliGemma 3B} \\
        \midrule
        Question & -0.034 & -0.034 & -0.074 & -0.044 \\
        Image    & 0.087  & 0.002  & -0.005 & -0.014 \\
        Context  & -0.036 & 0.016  & 0.045  & 0.026  \\
        \bottomrule
    \end{tabular}
    \label{tab:pcc_models}
\end{table}

\newpage
\section{Rebalancing Modality Importance for PaliGemma 3B} \label{apx:rebalance_pali}

\begin{figure}[ht!]
    \centering
    \includegraphics[page=11, trim=0cm 95cm 0cm  0cm, clip, width=\linewidth]{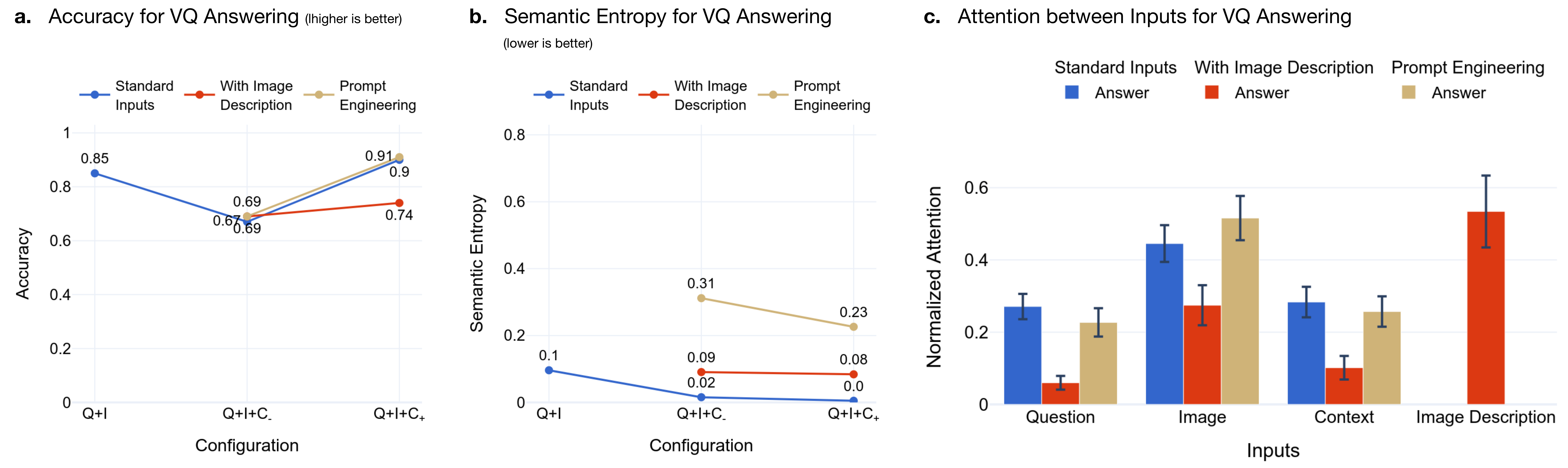}
 \caption{Two strategies to adjust modality importance: incorporating the image's textual description into the input and modifying the prompt to shift more attention toward the context. The impact of these changes is evaluated based on a. answer accuracy, b. model uncertainty, and c. attention attribution. The hereby experiments were conducted with the PaliGemma model.}
    \label{fig:vqa_rebalance_pali}
\end{figure}

\end{document}